\documentclass[times,review,nopreprintline]{elsarticle}

\usepackage{geometry}

\usepackage{tabularx}
\usepackage{blkarray}
\usepackage{soul}
\usepackage{framed,multirow}

\usepackage{amssymb}
\usepackage{latexsym}

\usepackage{url}
\usepackage{xcolor}

\usepackage{hyperref}

\usepackage{subcaption}
\usepackage{caption}
\usepackage{subcaption}

\usepackage{booktabs}

\definecolor{newcolor}{rgb}{.8,.349,.1}

\journal{Medical Image Analysis}

\usepackage{amsmath}

\usepackage{float}

\begin{document}

%\verso{Ozan Ciga \textit{et~al.}}

\begin{frontmatter}

\title{Learning to segment images with classification labels}%

\author[1]{Ozan Ciga\corref{cor1}}
\cortext[cor1]{Corresponding author: 
  e-mail: ozan.ciga@mail.utoronto.ca}
\author[1,2]{Anne L. Martel}

\address[1]{Department of Medical Biophysics, University of Toronto, Canada}
\address[2]{Physical Sciences, Sunnybrook Research Institute, Toronto, Canada}

%\received{}
%\finalform{}
%\accepted{}
%\availableonline{}
%\communicated{}

\begin{abstract}
%%%
Two of the most common tasks in medical imaging are classification and segmentation. Either task requires labeled data annotated by experts, which is scarce and expensive to collect. Annotating data for segmentation is generally considered to be more laborious as the annotator has to draw around the boundaries of regions of interest, as opposed to assigning image patches a class label. Furthermore, in tasks such as breast cancer histopathology, any realistic clinical application often includes working with whole slide images, whereas most publicly available training data are in the form of image patches, which are given a class label. We propose an architecture that can alleviate the requirements for segmentation-level ground truth by making use of image-level labels to reduce the amount of time spent on data curation. In addition, this architecture can help unlock the potential of previously acquired image-level datasets on segmentation tasks by annotating a small number of regions of interest. In our experiments, we show using only one segmentation-level annotation per class, we can achieve performance comparable to a fully annotated dataset.

%%%%
\end{abstract}

\begin{keyword}
%% MSC codes here, in the form: \MSC code \sep code
%% or \MSC[2008] code \sep code (2000 is the default)
% \MSC 41A05\sep 41A10\sep 65D05\sep 65D17
%% Keywords
% \KWD 
Weakly supervised learning, digital histopathology, whole slide images, image segmentation
\end{keyword}

\end{frontmatter}

%\linenumbers

%% main text

\begin{figure*}
         \centering
         \includegraphics[width=\textwidth]{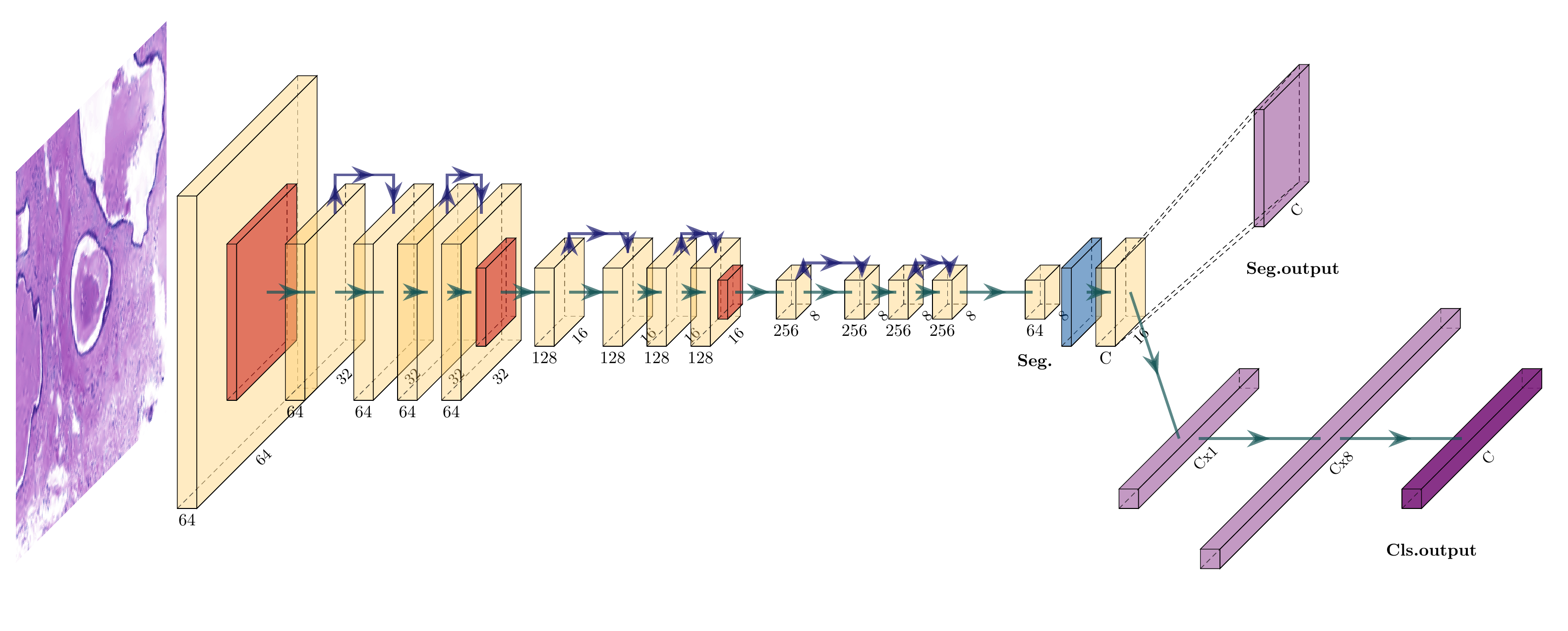}
     \caption{Proposed architecture. Blue arrows indicate residual connections of Resnet-18, the red squares are the maximum pooling operations, blue square is the unpooling (spatial up-sampling). After each convolutional layer (block) of the Resnet architecture, ReLU activation and batch-normalization \citep{ioffe2015batch} is applied. For details on the networks after Resnet-18 encoding (starting from the caption ``\textbf{Seg.}"), see tables \ref{tab:segmentation_module} and \ref{tab:classification_module}. }\label{fig:proposed_architecture_full}
\end{figure*}

\section{Introduction}

Labeled (i.e., annotated) data is critical to the performance of most machine learning approaches. Automated image analysis tasks require vast amounts of annotated data, where the annotation process is laborious and expensive. In tasks involving natural scenes and simple categorization of common objects (e.g., comparison of cats and dogs in ILSVR challenge \citep{imagenet_cvpr09}), annotation efforts can be crowdsourced \citep{10.1007/978-3-642-35142-6_14}. However, due to the complexity of medical imaging data as well as regulations limiting data sharing, crowdsourcing medical imaging tasks is more challenging \citep{rting2019ASO}, and tasks such as breast cancer classification require expert annotations. 

In medical image analysis, image annotation can refer to either patch level annotations, where an image patch of a certain size is given a single discrete class label (e.g., cancer or no cancer), or a segmentation mask, where each pixel in an image of an arbitrary size is given a label. While both need to be performed by a trained expert, generating fine boundaries to delineate regions of interest is much more time-consuming. We verified the time discrepancy between tasks by recording the time required for annotating metastases in lymph nodes for breast cancer. An expert pathologist initially identified regions that included the dominant type of cancer (classification), and then delineated the boundaries (segmentation). About 20\% of the examination time was spent on classification, whereas 80\% was spent on carefully drawing boundaries. For this task, about 22 minutes on average was spent on identifying the cancer. Depending on the size and the required detail for the region, pixel-level annotation times ranged from 15 minutes to two hours.
%For this task, about 12 minutes on average was spent on identifying the cancer, and 10 minutes was spent on distinguishing between ductal carcinoma versus ductal hyperplasia. Depending on the size and the required detail for the region, pixel-level annotation times ranged from 15 minutes to two hours.

%10-15 min to identify the cancer on WSI. another 10-15 to identify dcis vs ductal hyperplasia. polygon draw another 15.. identify mitosis takes 2 hours. 
%learning curve, 
%takes time to think about the problem etc + still takes time 

% num seg dataset vs cls dataset
%GLAS 2015, CAMELYON16, CAMELYON17, BACH, CRAG 2019....165+400+1000+10+213=1788 divide by /5=358
% NCT-CRC-HE-100k, Post-NAT-BRCA, BreastPathQ 2018, BACH 2018, PCam 2018, CAMELYON17, BreakHis 2016, HER2 Scoring 2016, TUPAC 2016...821+86+7909+1000+327680+500+96+96+100000=438188 divide by/9=49k

% sum([33(cls), 8, 21(seg), 16, 9, 27, 12])=num papers
%\textcolor{red}{26\% of the published work on histopathological image analysis on machine learning in the past seven year was related to classification tasks, whereas 17\% was on segmentation \citep{srinidhi2019deep}. Among the public computational pathology datasets, 5 out of 22 (22\%) involved segmentation tasks whereas 9 (41\%) datasets involved a classification task. On average, a classification dataset contains 49 thousand images whereas a segmentation dataset contains 358.}
The challenges associated with acquiring pixel-level annotations has led to a discrepancy in number of available images between segmentation and classification datasets. To verify that image-level annotations are more common than pixel-level annotations, we reviewed 22 public computational pathology image datasets \citep{srinidhi2019deep}. Five out of 22 (22\%) involved segmentation tasks, whereas nine (41\%) datasets involved a classification task. Importantly, classification datasets contained 49 thousand images on average, whereas  segmentation datasets only contained 358. Furthermore, most segmentation datasets were composed of WSIs, where each WSI contains less than six distinctly annotated regions on average.

In segmentation tasks, it is not uncommon for inter-expert agreement to be quite low, whereas it is less likely for two experts to disagree on the dominant class observed in a patch. For the former, training data is assumed to be exhaustively annotated, i.e., each pixel on an image is assigned its correct class. In histopathology, however, errors at the pixel level are inevitable. For example in annotating slides for the task of ductal carcinoma in situ (DCIS) annotation, pathologists may include background pixels between adjacent ducts in the DCIS region as this makes annotation much faster, or they may misclassify regions of DCIS elsewhere on the WSI \citep{seth2018automated}. This can lead to training with incorrectly labeled data and, since the training algorithm receives conflicting information, generalization performance will be degraded.

Therefore, it is desirable to combine these two sets of data to expedite the data annotation, and to have more reliable ground truth. With the currently available neural network architectures, however, utilizing data from both patch and pixel level annotations to improve performance is not straightforward. Image level patches cannot be used to train a fully connected segmentation network and, although it is possible to convert segmentation masks to labelled image patches in order to use a classification network, valuable pixel level boundary information  will be ignored. At test time, classifying patches on a WSI and tiling them to form a segmentation mask will lead to overestimating the structure. As the training is done using rectangular patches, each region will tend to have rectangular shape, even when a sliding window is used to obtain finer boundaries.

We propose a method to train both a segmentation network (primary task) using a limited number of segmented WSIs, and a classification network (auxiliary task) using labelled image patches. Our method simultaneously trains for the primary and auxiliary tasks in order to force the network to learn features useful for each task. The aim is to utilize the rich information content in (easier to obtain) patch level images to generate a feature representation that is then used to ``decode" this representation into a segmentation mask. In order to prevent overfitting to either task, both tasks share a pathway (network parameters or weights) which encodes the feature representation. The pathway may encode additional useful information into the representation with the help of additional data that is used to train the auxiliary task, which may have not been possible with just the limited primary task dataset. We believe that this method can be useful for clinical and applied researchers by enabling the use of cheaply obtained classification patches in medical image analysis tasks. The method can be leveraged for annotating images more efficiently to decrease the time required for the annotation and to increase the number of labeled images with negligible loss in performance. The classification patches can come from either the same or a different dataset for a given task.

%lay summary: grade 3 reading ability. this will reduce the effort required for labeling. this allows dig path imgs annot more efficiently, increasing amount of label in these datasets. quicker, cheaper, easier. to reviewer: we think relatively easy to understand but added these 2-3 sentences..

\begin{figure*}[!t]
         \centering
         \includegraphics[width=\textwidth]{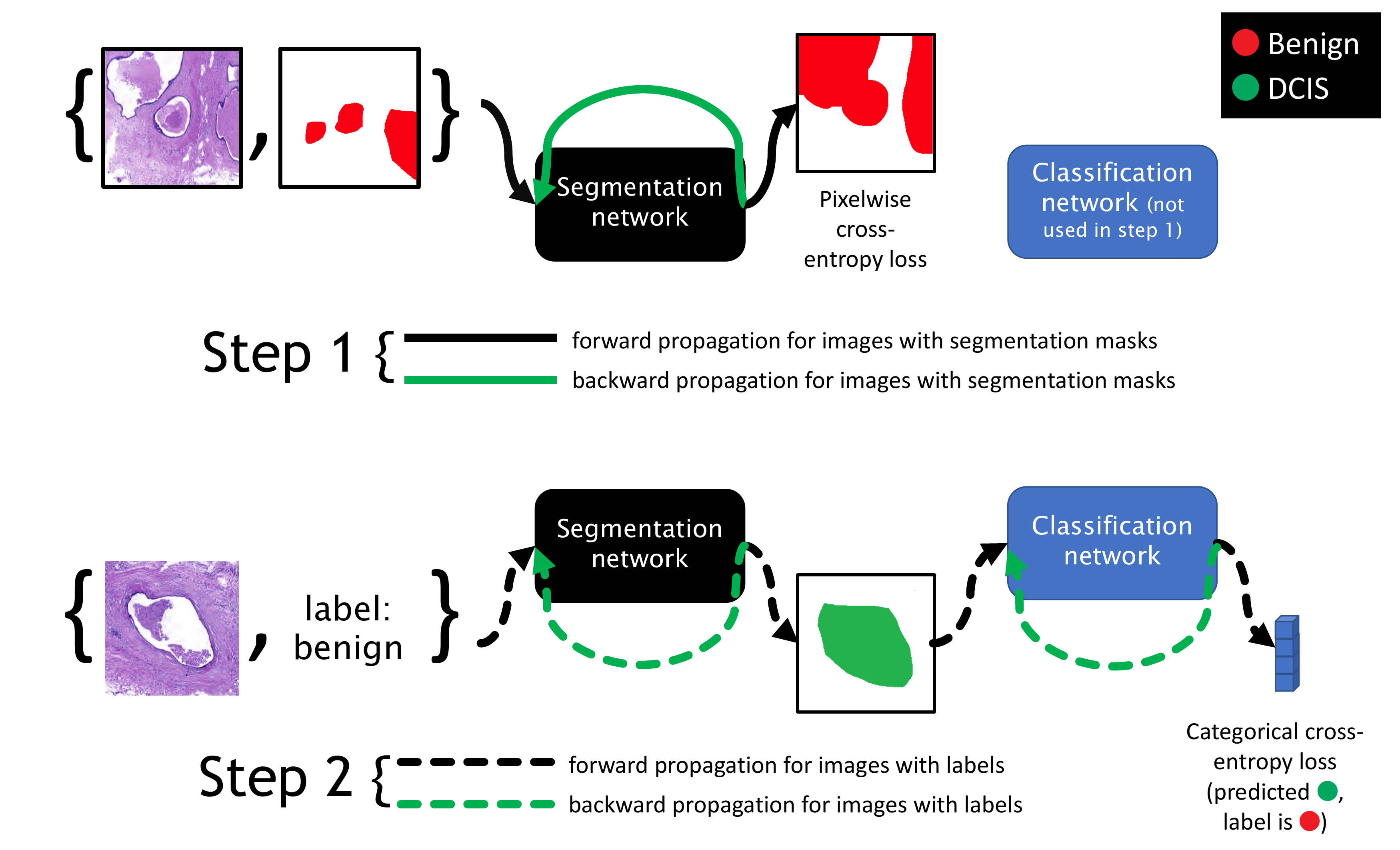}
     \caption{Overview of the training procedure for each batch. We leverage inductive biases present in U-Net architecture such as skip connections which lead to smoother segmentation boundaries. We modify the U-Net to make use of discrete labels by attaching a novel head at the segmentation output for transforming it into a predicted probability value. Then, we use the standard cross-entropy loss for training the network to improve its segmentation quality. Importantly, as the loss back propagates to earlier layers, which are mainly responsible for edge and texture detection to separate the structure from the background, it will diminish. In contrast, final layers that are responsible for high-level abstract information such as class type are modified more aggressively which corrects errors related to class predictions.}\label{fig:visual_abstract}
\end{figure*}

\subsection{Related work}

Unsupervised, self-supervised, and weakly supervised learning methods that attempt to reduce the burden of annotation by incorporating unlabeled data have been proposed previously.

Unsupervised learning refers to categorizing unlabeled data without supervision to form clusters that correlate with the desired task objective (e.g., clustering chest X-rays into tuberculosis vs. healthy). \cite{AHN2019140} proposed a hierarchical unsupervised feature learning method using a sparse convolutional kernel network to identify invariant characteristics on X-ray images by leveraging the sparsity inherent to medical image data. In histopathological image analysis, sparse autoencoders have been utilized for unsupervised nuclei detection \citep{xu2015stacked, hou2019sparse}, and for more complex tasks such as cell-level classification, nuclei segmentation, and cell counting, Generative Adversarial Networks (GANs) have also been employed \citep{hu2018unsupervised}.

In self-supervised learning, the aim is to use the raw input data to generate artificial and cost-free ground truth labels in order to train a network, which can then be used as initialization for training on a separate task with limited data. \cite{noroozi2016unsupervised, gidaris2018unsupervised} exploited the spatial ordering of a natural scene image to generate labels. This form of training was also adopted in medical image analysis, where \cite{taleb2019multimodal} spatially reordered multi-organ MR images to train an auxiliary reordering self-supervision network, and used it to train a network for tasks such as brain tumor segmentation with limited data. \cite{5b38f8cb124741559716f8dc05e36932} utilized image reconstruction, rotations, and grayscale to RGB colorization schemes as supervision signal for chest computed tomography images. \cite{CHEN2019101539} used context restoration of images from different modalities including 2D ultrasound images, abdominal organs in CT images, and brain tumours in multi-modal MR images. 

We consider techniques where labeled data is insufficient in amount, or where labels are noisy or inaccurate, or where annotations do not directly correspond to the task at hand (e.g., using coarse, image-level annotations to train a semantic segmentation network) under weakly supervised learning. \cite{Li2019weakly, Qu2019weakly} trained models with sparse sets of annotations, consisting only single-pixel annotations to perform mitosis segmentation, whereas \citep{Yang2018boxnet} utilized a coarse bounding box to perform gland segmentation in histology images. Similarly, \citep{Bokhorst19} utilized a U-Net \citep{10.1007/978-3-319-24574-4_28} and sparsely annotated various structures of interest in colorectal cancer patient WSIs to achieve semantic segmentation. In an effort to combine classification with segmentation, \cite{mehta2018ynet} trained a network where they tiled patch level classification outputs with the segmentation outputs to refine the segmentation output. \cite{WONG2018105} trained a U-Net segmentation network, and used its encoder's features to fine-tune a convolutional network on a nine-class cardiac classification task that showed high performance in a low data setting, compared to a network trained from scratch. \cite{rajchl2017employing} propose an outlier removal method for liver computed tomography segmentation problem which relies on sequentially averaging the ground truth bounding boxes from multiple annotators. At each step, regions that had high disagreement with the ``consensus" are removed from the consensus ground truth, and a more refined 3D segmentation mask is obtained. \cite{remez2018learning} follows an adversarial approach to achieve segmentation using weak annotations in the form of bounding boxes. Specifically, they use a generator network to predict the semantic segmentation mask of a given object within the bounding box. Given a coarse segmentation mask, they combine (blend) the masked object with a separate background. The discriminator network is asked to compare the original object within the bounding box with the combined image. Simultaneous adversarial training leads to improved semantic segmentation since discriminator is only deceived if the combined image looks realistic, which requires the segmentation outline to be accurate for realistic blending of the background with the object.

Current unsupervised methods are not capable of processing complex images, and generally cannot be applied to images larger than $64\times64$ pixels, whereas in histopathological images dimensions are much larger. Similarly, most of the existing self-supervised techniques are not applicable to histopathological images, since structures in these images are elastic and may form infinitely many valid groupings (e.g., a fat cell can be next to, above, below, or be surrounded by stroma). In contrast, weakly supervised learning methods we have reviewed have been previously successful in various digital histopathology tasks including classification and segmentation, and are capable of working with larger images sizes. Therefore in this work, we focus on a method that is based on weakly supervised learning, as opposed to tackling segmentation with limited data with unsupervised or self-supervised techniques.

\subsection{Contributions}

In this work, we propose a simple architecture to alleviate the burden on the annotator by combining data acquired from two different processes, either by classification labels on patches, or segmentation masks from either whole slide images or image patches. We consider our method as a form of weakly supervised learning where the training data includes different levels of information, namely on patch- and pixel-level. We believe our work is clinically relevant and can help expedite the process of data acquisition to bridge the gap between advancements in the deep learning which heavily rely on data with clinical applicability. Our contributions are listed as follows.

\begin{itemize}
    \item A simple modification to the existing Resnet architecture to perform segmentation and classification simultaneously, while leveraging easier-to-label classification patches to improve segmentation performance with small amounts of labeled segmentation data,
    \item Two data preprocessing techniques that aim to alleviate the class imbalance problem (mainly due to the dominant background) observed in whole slide images in digital pathology: a simple background thresholding procedure in HSV color space, and a method to extract ground truth segmentation mask more efficiently for faster and more reliable training performance.  
\end{itemize}

\section{Methods}

\subsection{Data processing}

We propose two data preprocessing techniques that can be used in segmentation tasks on WSIs. Our aim is to remove healthy tissue that is not relevant to tasks such as cancer segmentation by thresholding, and to alleviate class imbalance observed in WSIs, using a novel ground truth extraction technique. For further details, see \ref{apx:data_preprocess}.

\subsection{Architecture}\label{sec:architecture}

Our architecture is summarized in Fig. \ref{fig:proposed_architecture_full}.  We use the third layer output (out of a total of four layers) of the Resnet-18 architecture \citep{he2016deep} for encoding input images, and use architectures given in tables \ref{tab:segmentation_module} and \ref{tab:classification_module} for decoding (i.e., generating segmentation mask) and classification, respectively. Note that the classification path is \textit{going through} the segmentation network. This was made to first obtain a segmentation output for the classification images, and use this prediction to infer the classification label. We reason that segmentation network should be partially capable of segmenting structures even in the low data settings. Then, a transformation network (the classification module) is used to aggregate the information in this prediction to infer a class. Using the image level training data, the network is trained to discard or modify incorrect class assignments in segmentation outputs while keeping and reinforcing correct assignments, by modifying the feature representation of input images. Importantly, since segmentation implicitly contains the region's label, classification is unavoidable. Furthermore, delineating boundaries for a region is mostly an edge detection task, which can be encoded in low-level features from a few samples, whereas more samples are required to distinguish between different types of disease. 

An overview is given in Fig. \ref{fig:visual_abstract}. For each batch of training, we perform two alternating steps using pixel-level images (with segmentation masks) and image-level labels (images with only classification labels) in the batch: In step 1, pixel level images are used to train the network with input images and segmentation masks with the standard backpropagation algorithm on the segmentation network (encoder+decoder) without passing through the classification layer. In step 2, the data with only image level labels are passed through the segmentation network to obtain a segmentation mask output, which is then transformed by the classification layer to obtain the classification output vector $\in \mathbb{R}^C$, where $C$ is the number of classes. This vector is used for backpropagation with cross entropy loss as an error signal to update segmentation network weights to correct the segmentation mask for the given image. For instance, the bottom part (step 2) of Fig. \ref{fig:visual_abstract} depicts the case where the network incorrectly assigns \textit{DCIS} (green) to the large region. Given the ground truth label is \textit{benign}, this assignment is invalid, and the errors are corrected during training with backpropagation.

% The middle part in Fig. \ref{fig:visual_abstract} depicts the case where the network incorrectly assigns \textit{DCIS} (green) to the large region. Given the ground truth label is \textit{benign}, this assignment is invalid, and the errors are corrected during training with backpropagation. % However, it should be noted that the proposed method will not be able to correct errors if the incorrect annotations are of the same class as the image level label for the given patch. For instance, if the two small regions were labeled as \textit{benign}, the network would reinforce its prediction, as opposed to correctly identifying these regions as background. In our experiments, we observed that the probability of making this type of error decreases as the training set size increases. In addition, we observed that the data collection process is crucial to the success of the pipeline. The model is able to distinguish between the background and the foreground given only a few segmentation patches, by learning low-level features (e.g., edge or blob extractors) to outline the boundaries of the foreground structure. However, the correct class assignments of the segmented regions are more challenging compared to the background and foreground separation. Since the error signal from the classification layer does not specify which pixels should be corrected, image level data should be collected where the majority of the pixels on the patch are either background or the foreground class assigned to that patch.

\subsection{Implementation details}

We use the Adam optimizer with $\beta_1 = 0.9$, $\beta_2 = 0.999$, learning rate of 0.0001, batch size of 20, and weighted cross entropy loss function with weights proportional to the pixelwise class distributions among training images. Cross entropy loss is applied pixelwise for segmentation, and per item for classification. We use stain normalization \citep{macenko2009method} based on a single reference image selected from the training set, and simultaneously optimize classification loss $\mathcal{L}_{cls}$ and segmentation loss $\mathcal{L}_{seg}$ by minimizing the quantity $ \mathcal{L}_{cls} + \mathcal{L}_{seg}$. 

We train on a single NVIDIA GeForce GTX 1080 GPU with 8 GB graphical memory, 32 GB of physical memory, and an Intel(R) Core(TM) i7-8700 CPU @ 3.20GHz. 100 epochs of the training takes 6.2 minutes for a dataset with 1015 RGB images, with a patch size of $128\times128$ pixels. The ratio of segmentation and classification patches have negligible effect on the training duration. The source code for our method is available at \url{https://github.com/ozanciga/learning-to-segment}.

\section{Experiments}

\subsection{Data}

\begin{figure*}
    \centering
    \begin{subfigure}[b]{0.32\textwidth}
        \centering
         \includegraphics[width=0.49\textwidth]{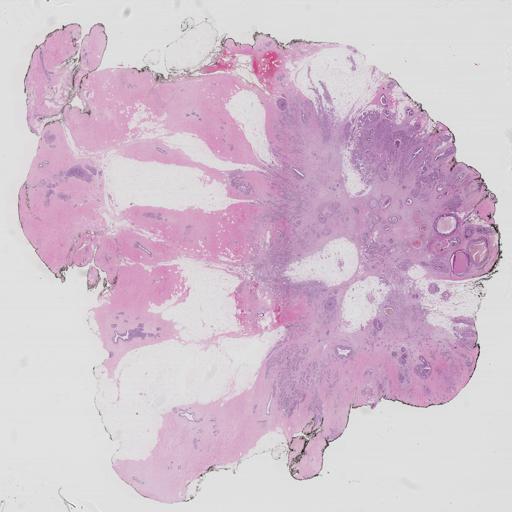}
         \includegraphics[width=0.49\textwidth]{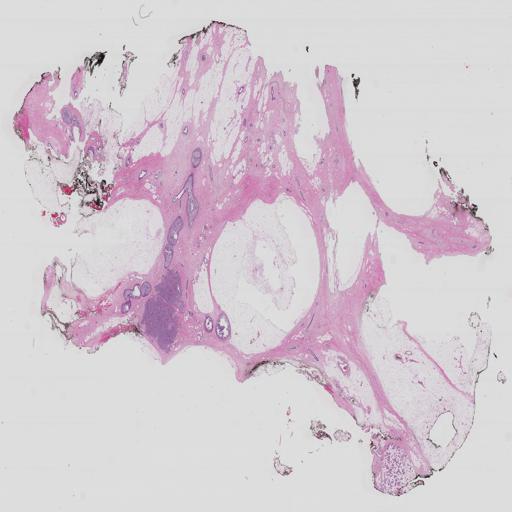}
        \caption{ICIAR BACH 2018}
    \end{subfigure}\hfill
    \begin{subfigure}[b]{0.32\textwidth}
        \centering
         \includegraphics[width=0.49\textwidth]{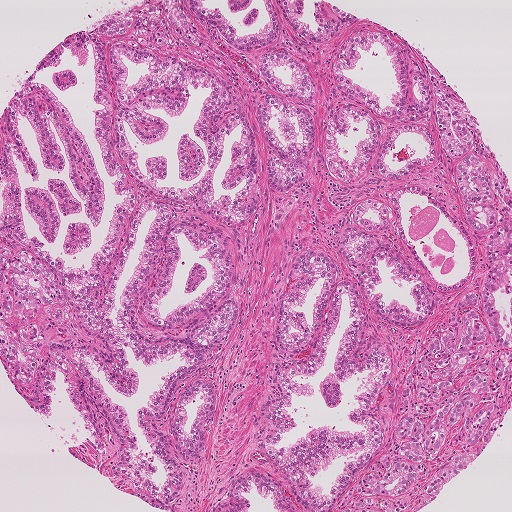}
         \includegraphics[width=0.49\textwidth]{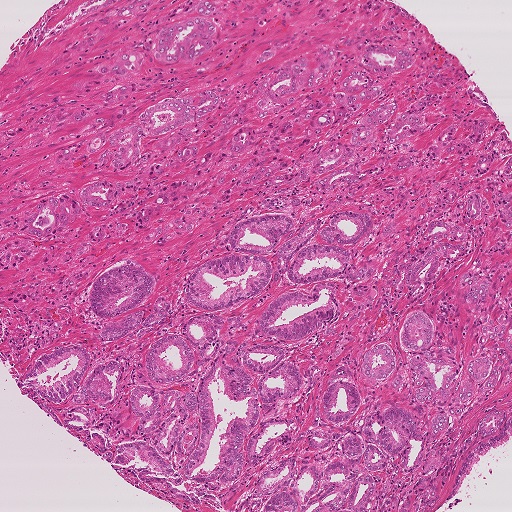}
        \caption{Gleason2019}
    \end{subfigure}\hfill
    \begin{subfigure}[b]{0.32\textwidth}
        \centering
        \includegraphics[width=0.49\textwidth]{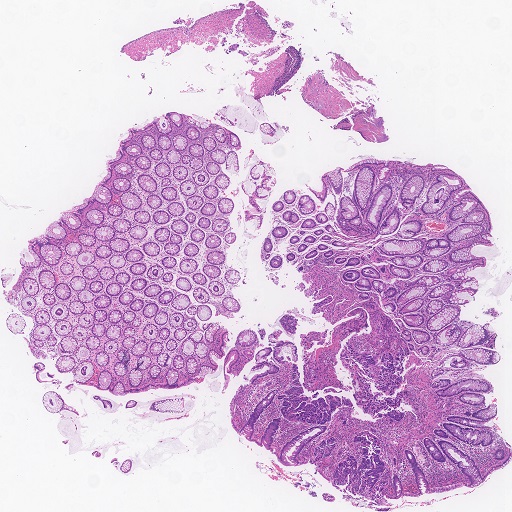}
         \includegraphics[width=0.49\textwidth]{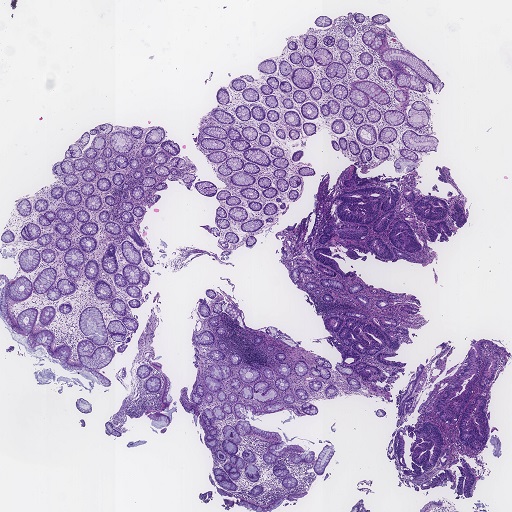}
        \caption{DigestPath2019}
    \end{subfigure}
    \caption{Sample images from each dataset.}\label{fig:dataset_sample_images}
\end{figure*}

We demonstrate the effectiveness of our method in a variety of clinically relevant digital histopathology tasks by evaluating it on three separate public datasets from different organs (prostate, colon and breast). The datasets detailed below are open access, and research ethics board approvals for each dataset were individually cleared. Sample images from each dataset are provided in Fig. \ref{fig:dataset_sample_images}.

\paragraph{ICIAR BACH 2018: Breast cancer histology} BreAst Cancer Histology images (BACH) dataset is a grand challenge in breast cancer classification and segmentation \citep{aresta2019bach}, where the dataset is composed of both microscopy and whole slide images. The challenge is split into two parts, A and B. For part A, the aim is to classify each microscopy image into four classes, normal tissue, benign tissue, ductal carcinoma in situ (DCIS), and invasive carcinoma, whereas in part B, the task is to predict the pixelwise labeling of WSI into same four classes, i.e., the segmentation of the WSI. The classes are mapped to labels 0, 1, 2, 3 for the classification task, and white, red, green, blue in the following figures, representing normal, benign, DCIS, and invasive classes in both tasks, respectively. The dataset consists of 400 training and 100 test microscopy images, where each image has a single label, and 20 labeled WSIs with segmentation masks (split into 10 training and 10 testing images), in addition to 20 unlabeled WSIs with possible pathological sites.

\paragraph{Gleason2019: Grading of prostate cancer} Gleason score, ranging from 1 (healthy) to 5 (abnormal), is a strong prognostic predictor for the prostate cancer, and is used for assessing the grade from a biopsy image. Gleason2019 challenge consists of 244 tissue micro-array (TMA) images and their corresponding pixel-level annotations detailing the Gleason grade of each region on an image. Each image is separately annotated by four to six experts, which are called maps. For our experiments, we use the first map as our ground truth and predict the Gleason grade (1 to 5) of each pixel on a TMA.

\paragraph{DigestPath2019: Colonoscopy tissue segmentation} The aim of DigestPath2019 is to identify the early stage colon tumors from small tissue slices \citep{li2019signet}. The dataset consists of 660 image patches with binary pixel-level masks (benign and malignant) from 324 WSIs scanned at 20$\times$ resolution. The average size of each image patch is of $5000 \times 5000$ pixels, which are resized to $1024 \times 1024$ for our experiments.

\subsection{Metrics}

We report the metrics defined by the Equations \ref{eqn:precision}-\ref{eqn:f1}. We report two variants of F1 scores called the macro and micro F1. Both metrics are calculated class-wise, and the macro weighs each class-wise score equally whereas the micro considers the class imbalance, weighing the scores per ground truth ratios on the WSI. These metrics are equivalent when the F1 measure is calculated on a two-class problem.

\begin{align}
  &\frac{tp}{tp+fp} &&\text{precision}\label{eqn:precision}\\
  &\frac{tp}{tp+fn} &&\text{recall}\label{eqn:recall}\\
  & \frac{precision\cdot recall}{(precision+recall)/2} && \text{F1 score}\label{eqn:f1}\\
\end{align}

\subsection{Experimental setting}

 \begin{figure}[h]
     \centering
     \includegraphics[width=0.225\textwidth]{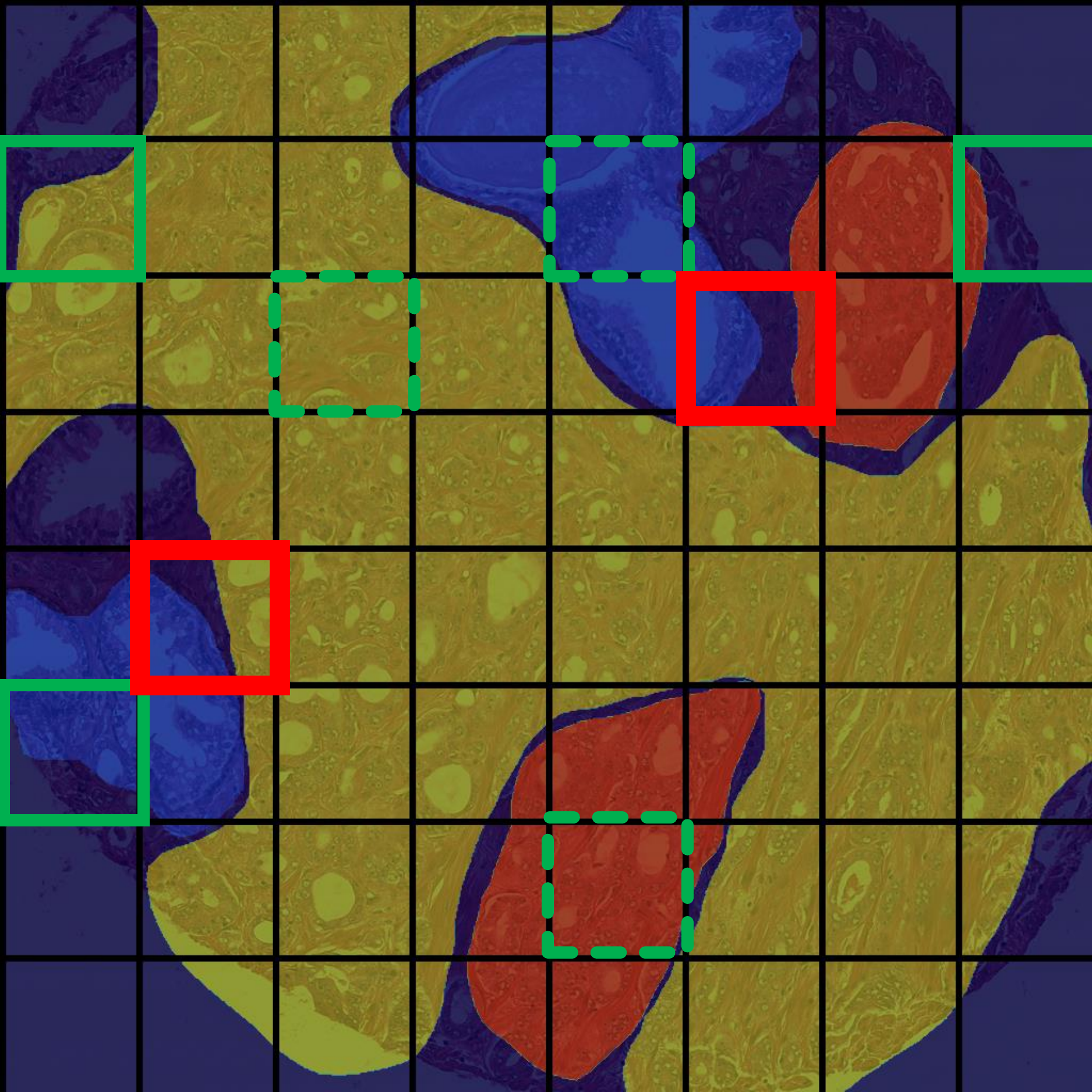}
     \caption{Tile extraction for DigestPath2019 and Gleason2019. The tiles outlined with solid green boxes represent the segmentation patches, dashed green boxes represent the classification patches, and the tiles in red boxes are ignored as they do not belong to either category.}
     \label{fig:tile_extract}
 \end{figure}

For experiments with the ICIAR BACH 2018 dataset, we use 3 WSIs from the part B (converted into segmentation patches) and the complete dataset from the part A (classification patches) of the BACH challenge for training, 1 WSI that includes all four classes as the validation set and evaluate our method on the remaining 6 WSIs. For experiments with Gleason2019 and DigestPath2019, we use 50\% of the dataset as our training set, 25\% as the validation, and the remaining 25\% as the test set. Since both of these datasets only include images and their corresponding segmentation masks, we use the following procedure to obtain classification patches, as visualized in Fig. \ref{fig:tile_extract}. We split each segmentation mask into tiles of size $128 \times 128$ pixels. If a dominant class is covering $\geq 90\%$ of the tile, then it is considered as a classification patch (Fig. \ref{fig:tile_extract}, dashed green boxes). A tile that only contains two classes is considered as a segmentation patch (Fig. \ref{fig:tile_extract}, solid green boxes), and any other tile is ignored (e.g., tiles with three unique classes, Fig. \ref{fig:tile_extract}, red boxes). 

We conduct a total of 195 experiments, each run for 100 epochs. We use $s\%$ of segmentation patches for \textbf{S} (only segmentation), and $(100-s)\%$ of classification patches in the case of \textbf{S+C} (segmentation + classification) experiments, where $s \in \{ 0, 1, 2.5, 5, 7.5, 10, 15, 20, 25, 30, 40, 50, 75, 100 \}$. We also use 100\% classification patches and $s\%$ of segmentation patches in \textbf{S+C*} setting, to examine if the addition of classification patches improves or degrades the performance. For $s=0\%$, we only use classification patches, hence for the \textbf{S} setting, the network predicts random outputs. For \textbf{S+C} and \textbf{S+C*} settings, $s=0\%$ reduces to a classifier which predicts one value per patch. We run 5 iterations per percentage value, randomly picking the same $s\%$ of training data for \textbf{S}, \textbf{S+C}, and \textbf{S+C*}, each time to accurately reflect the performance for each setting. 

We conduct experiments to assess if the segmentation step improves the quality of learned representations for the classification task. We use up to 50\% of classification patches for each dataset for training our model as before, and evaluate the classification network's performance on the remaining 50\%. For evaluating the effect of segmentation patches on the classification performance, we modify our experimental settings in classification head experiments. We use 100-2c\% of segmentation patches and c\% of classification patches for the \textbf{$S_2+C_2$}, 100-2c\% of segmentation patches, and 50\% of classification patches for the \textbf{$S_2+C^*_2$}. For each dataset, in addition to \textbf{$S_2+C_2$} and \textbf{$S_2+C^*_2$}, we report the setting \textbf{$S^*_2+C_2$}, where we use 100\% of segmentation patches, and the number of classification patches is varied from 0 to 50\%. We summarize our results in tables \ref{tab:classification_head_results} and \ref{tab:classification_head_results_raw}, and Fig. \ref{fig:classification_head_results}.

\section{Results and Discussion}

\begin{center}
\begin{table}
\caption{Classification results using the classification head on 50\% of the classification patches.}\label{tab:classification_head_results}
\begin{tabular}{p{1.5cm}|p{1.25cm} p{1.25cm} p{3cm}}
Dataset & \# of images & Accuracy & Confusion matrix \\ 
\midrule
ICIAR BACH 2018 & 400 & 80\% & $\begin{pmatrix}
34 & 4 & 0 & 2 \\ 
6 & 35 & 0 & 1 \\ 
6 & 5 & 46 & 1 \\ 
4 & 6 & 4 & 46 
\end{pmatrix}$ \\
Gleason 2019 & 1774 & 65\% & $\begin{pmatrix}
53 & 0 & 1 & 29 & 21 \\
7 & 25 & 7 & 4 & 33 \\
2 & 0 & 36 & 19 & 11 \\
1 & 0 & 12 & 201 &  81 \\
2 & 0 & 15 & 59 & 268 \\
\end{pmatrix}$ \\
Digest Path 2019 & 1764 & 82\% & $\begin{pmatrix}
256 & 53 \\ 
100 & 473
\end{pmatrix}$
\\ \bottomrule
\end{tabular}
\end{table}
\end{center}

\begin{figure*}
     \centering
     \begin{subfigure}[b]{0.32\textwidth}
         \centering
         \includegraphics[width=\textwidth]{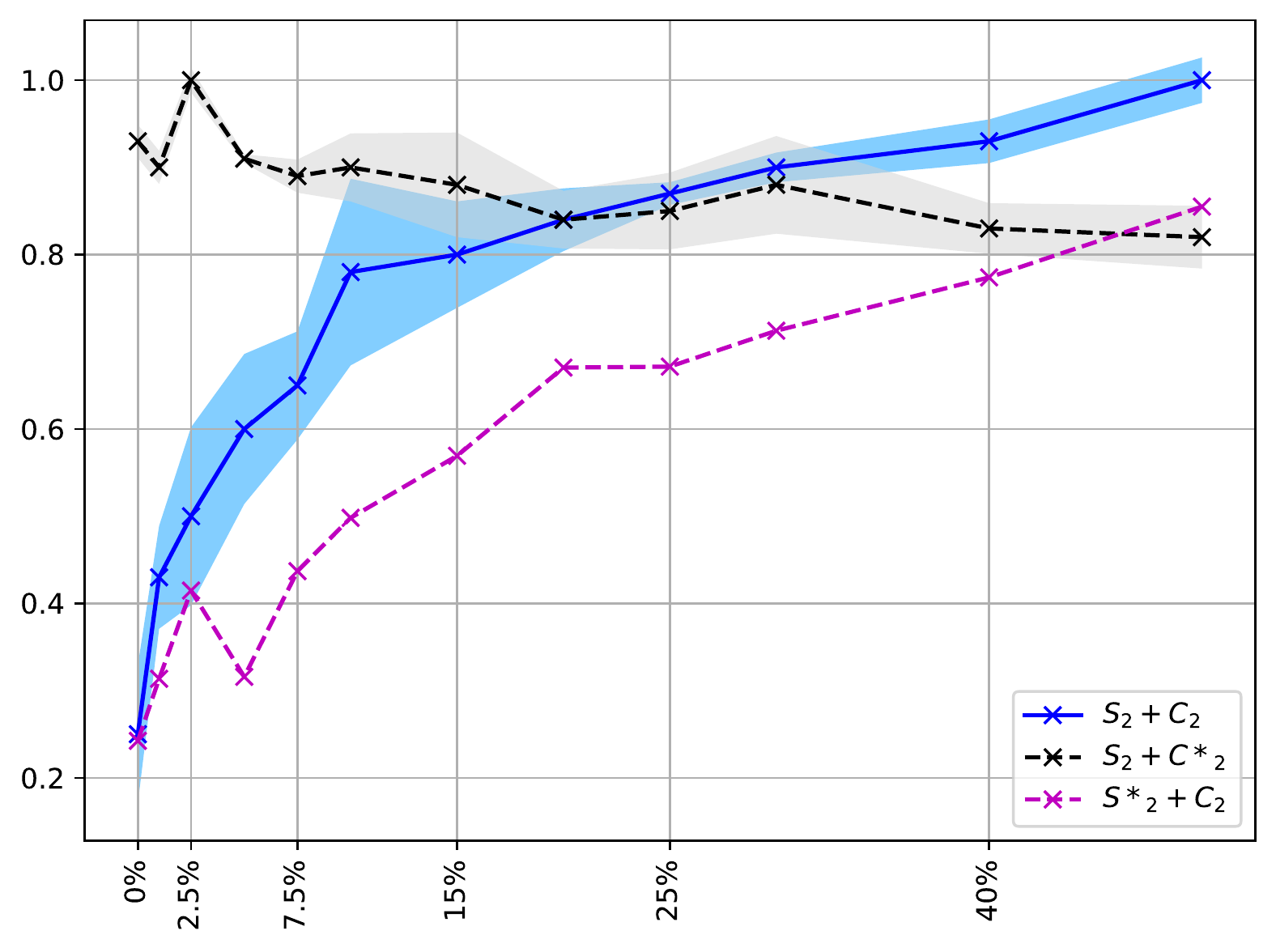}
         \caption{ICIAR BACH 2018}
     \end{subfigure}
     \begin{subfigure}[b]{0.32\textwidth}
         \centering
         \includegraphics[width=\textwidth]{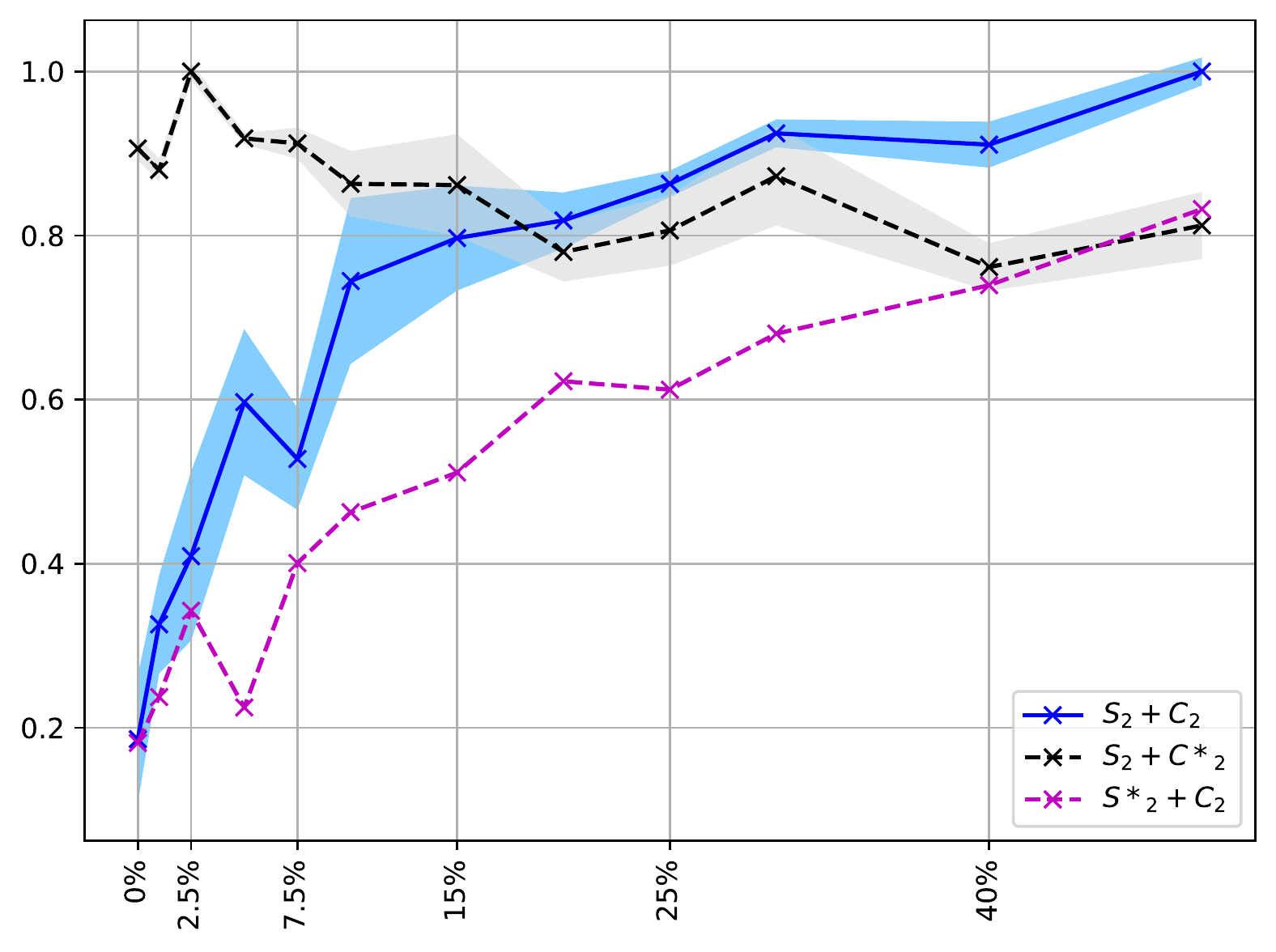}
         \caption{Gleason2019}
     \end{subfigure}
      \begin{subfigure}[b]{0.32\textwidth}
     \centering
     \includegraphics[width=\textwidth]{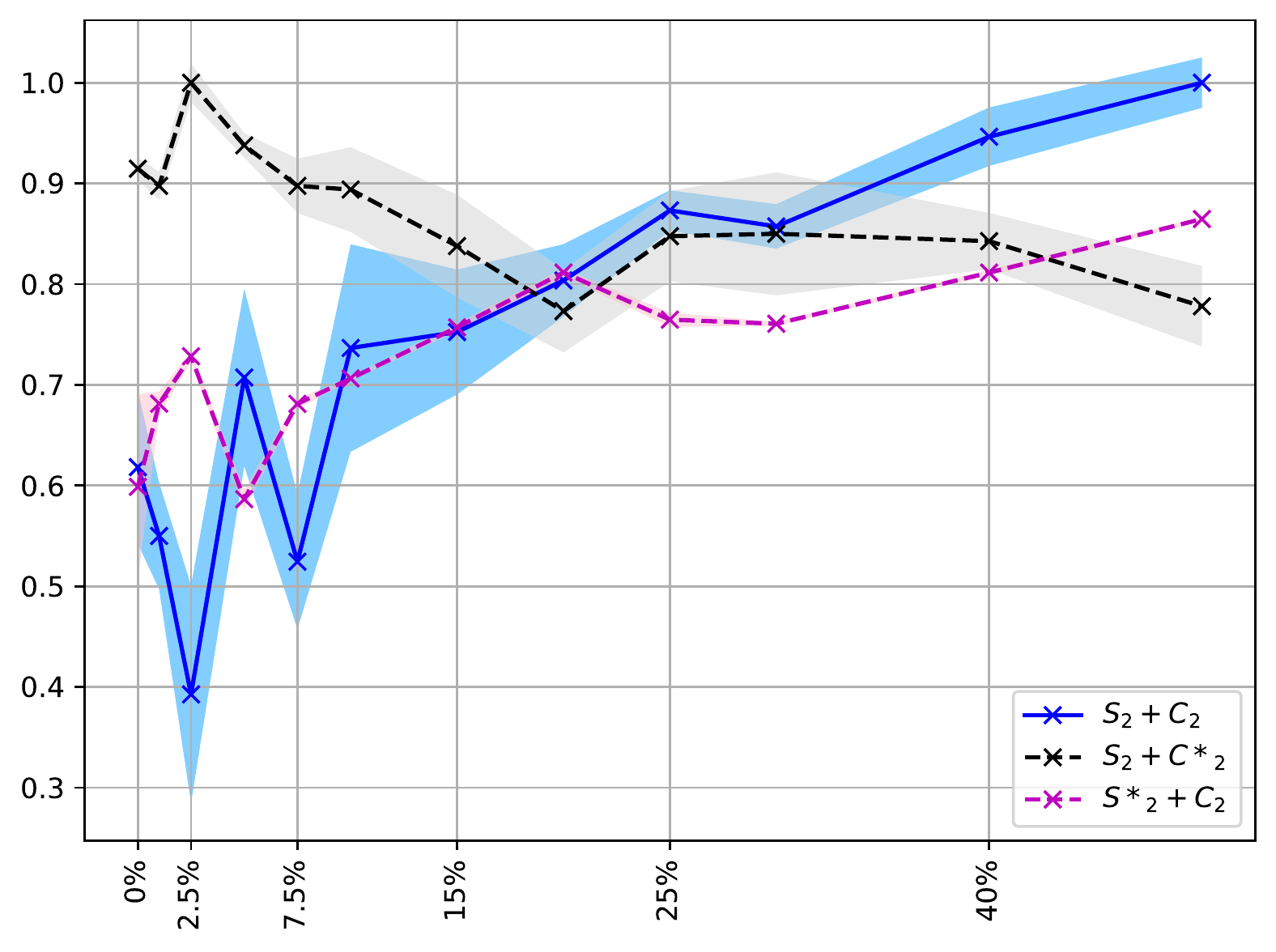}
     \caption{DigestPath2019}
 \end{subfigure}
        \caption{Accuracy results for the classification task for the three datasets. These results are normalized to c=50\%. For the raw results, please refer to \ref{tab:classification_head_results_raw}.}
        \label{fig:classification_head_results}
\end{figure*}

In order to assess if our network is useful for classification tasks, we conduct validation experiments on classification patches. We observe that segmentation task does not improve the classification task's performance on any dataset (Fig. \ref{fig:classification_head_results}, and Table \ref{tab:segmentation_results_raw} for the raw metrics). While the classification performance is positively correlated with the number of training samples, the addition of segmentation patches negatively impacts the performance. This adverse effect is observable in Fig. \ref{fig:classification_head_results} where for \textbf{$S_2+C^*_2$} and \textbf{$S^*_2+C_2$}, the addition of segmentation patches decrease the accuracy. Specifically, we achieve the best performance when we use 0\% of segmentation, and 50\% of classification patches and any addition of segmentation patches decrease the performance. Therefore, we conclude features obtained by training a network for segmentation are not useful for classification.

\begin{figure*}
     \centering
     \begin{subfigure}[b]{0.4\textwidth}
         \centering
         \includegraphics[width=\textwidth]{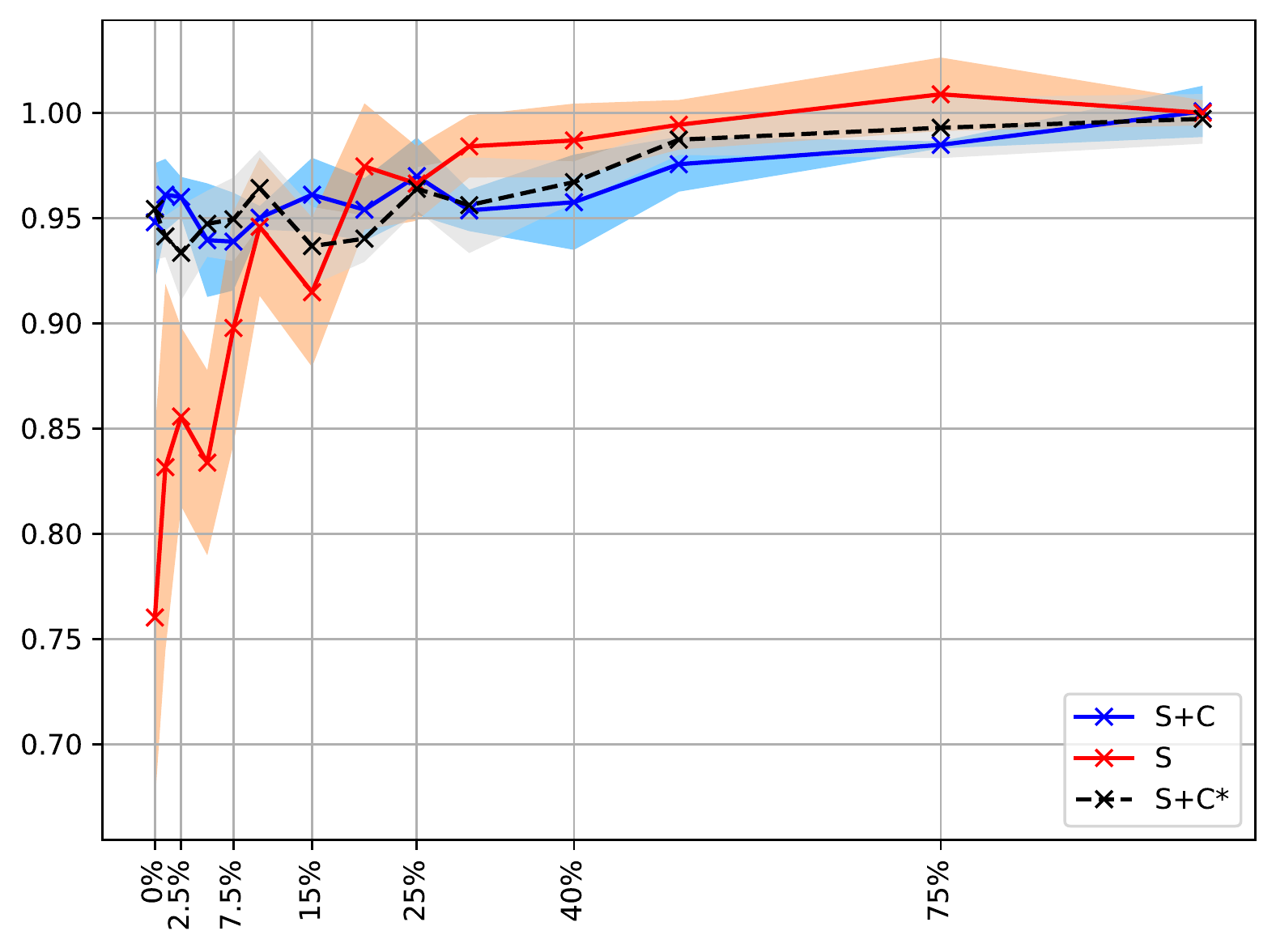}
         \caption{F1 (macro), ICIAR BACH 2018}
         \label{fig:c_s_f1mac_bach}
     \end{subfigure}
     \begin{subfigure}[b]{0.4\textwidth}
         \centering
         \includegraphics[width=\textwidth]{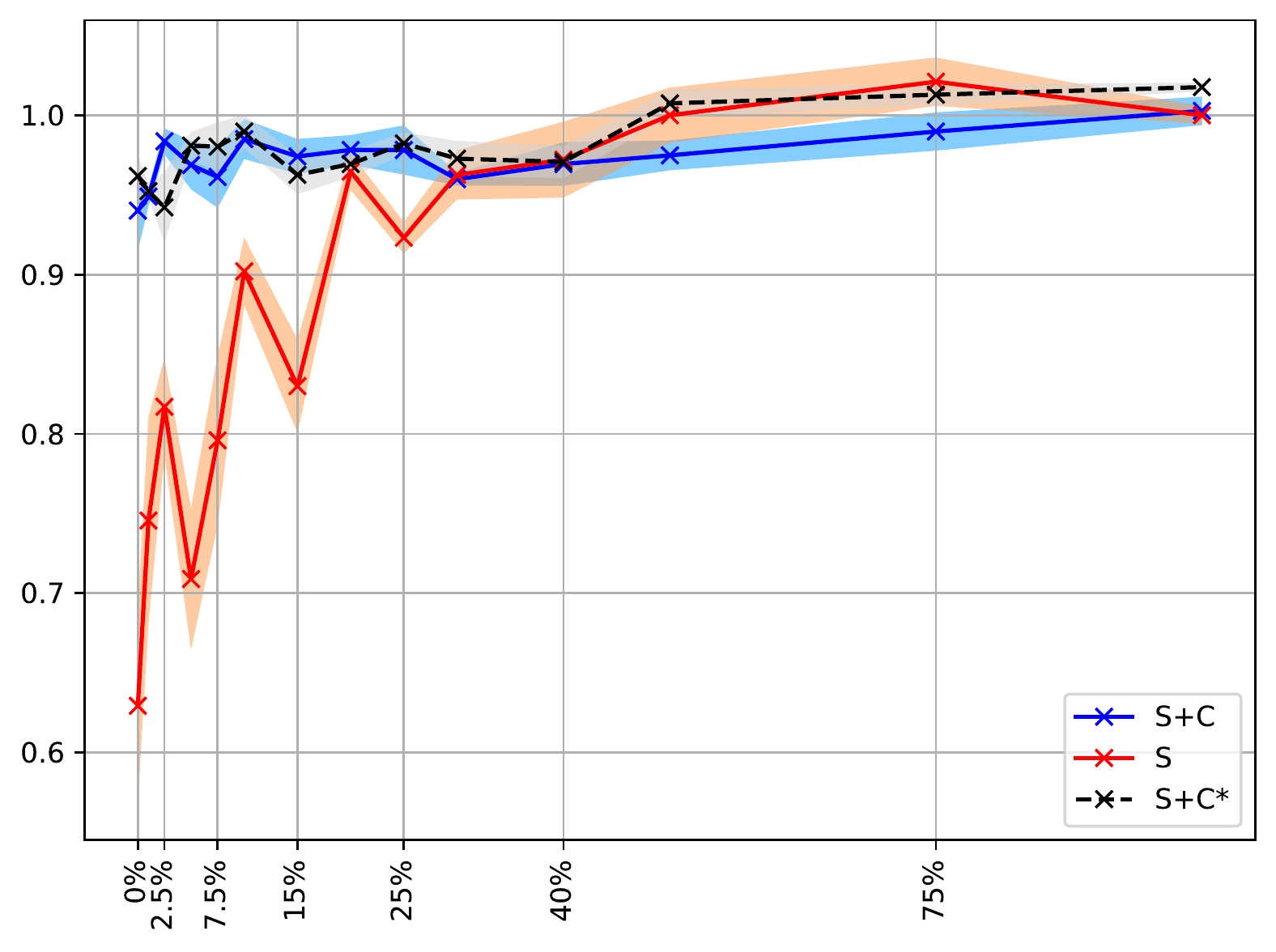}
         \caption{F1 (micro), ICIAR BACH 2018}
         \label{fig:c_s_f1mic_bach}
     \end{subfigure}
    \hfill
     \begin{subfigure}[b]{0.4\textwidth}
         \centering
         \includegraphics[width=\textwidth]{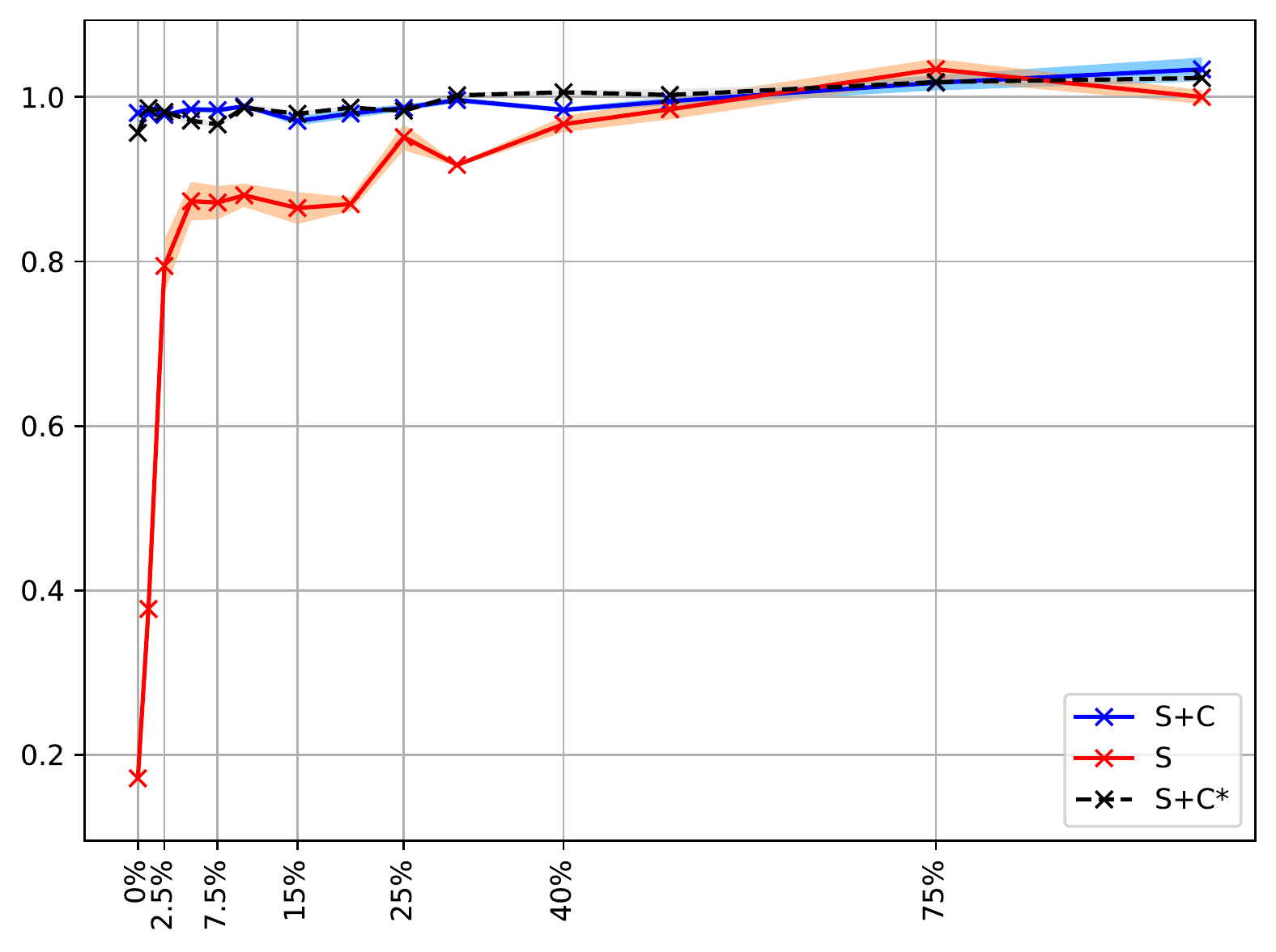}
         \caption{F1 (macro), Gleason 2019}
         \label{fig:c_s_f1mac_gleason2019}
     \end{subfigure}
     \begin{subfigure}[b]{0.4\textwidth}
         \centering
         \includegraphics[width=\textwidth]{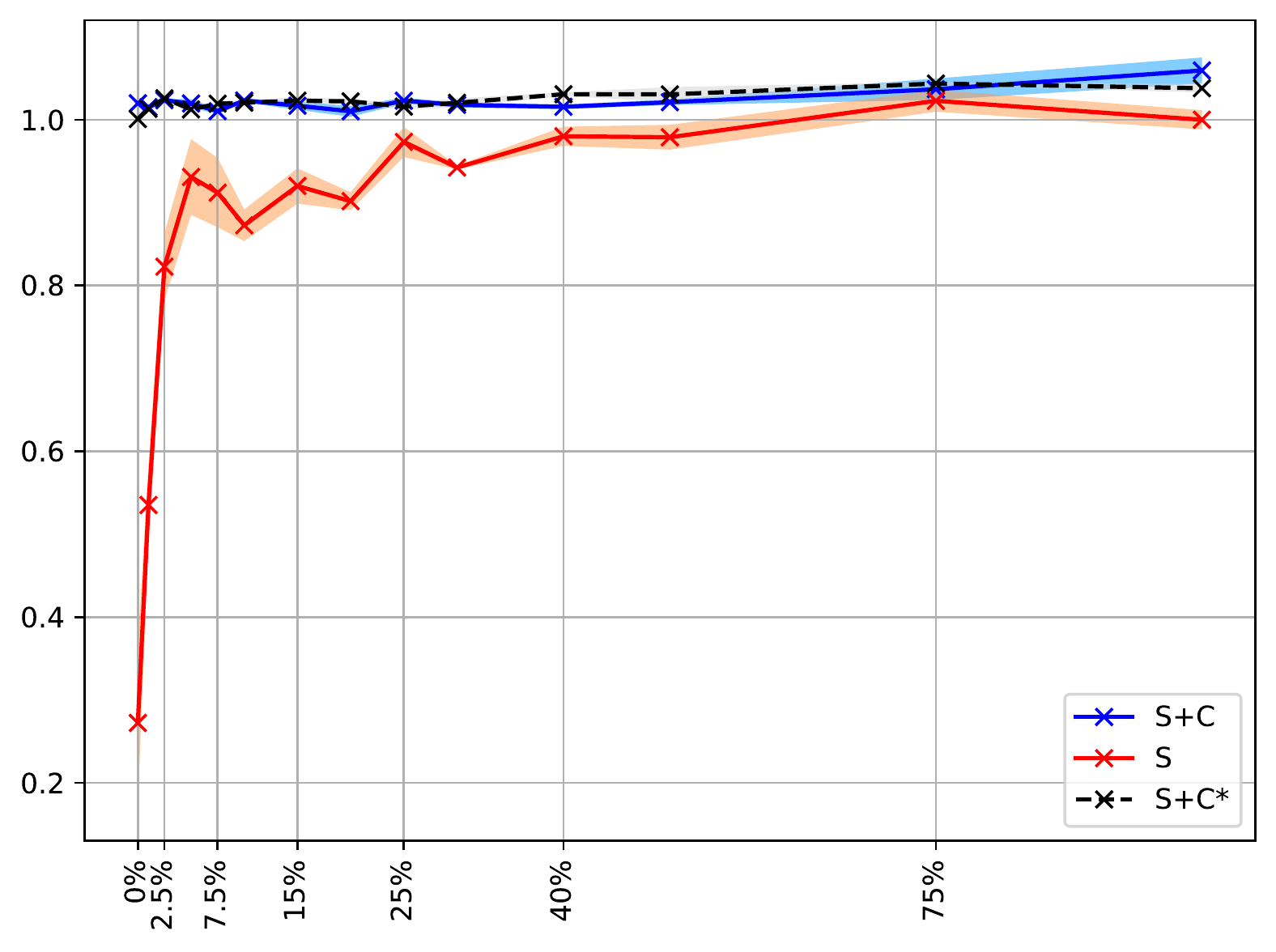}
         \caption{F1 (micro), Gleason 2019}
         \label{fig:c_s_f1mic_gleason2019}
     \end{subfigure}
     \vfill
     \begin{subfigure}[b]{0.4\textwidth}
         \centering
         \includegraphics[width=\textwidth]{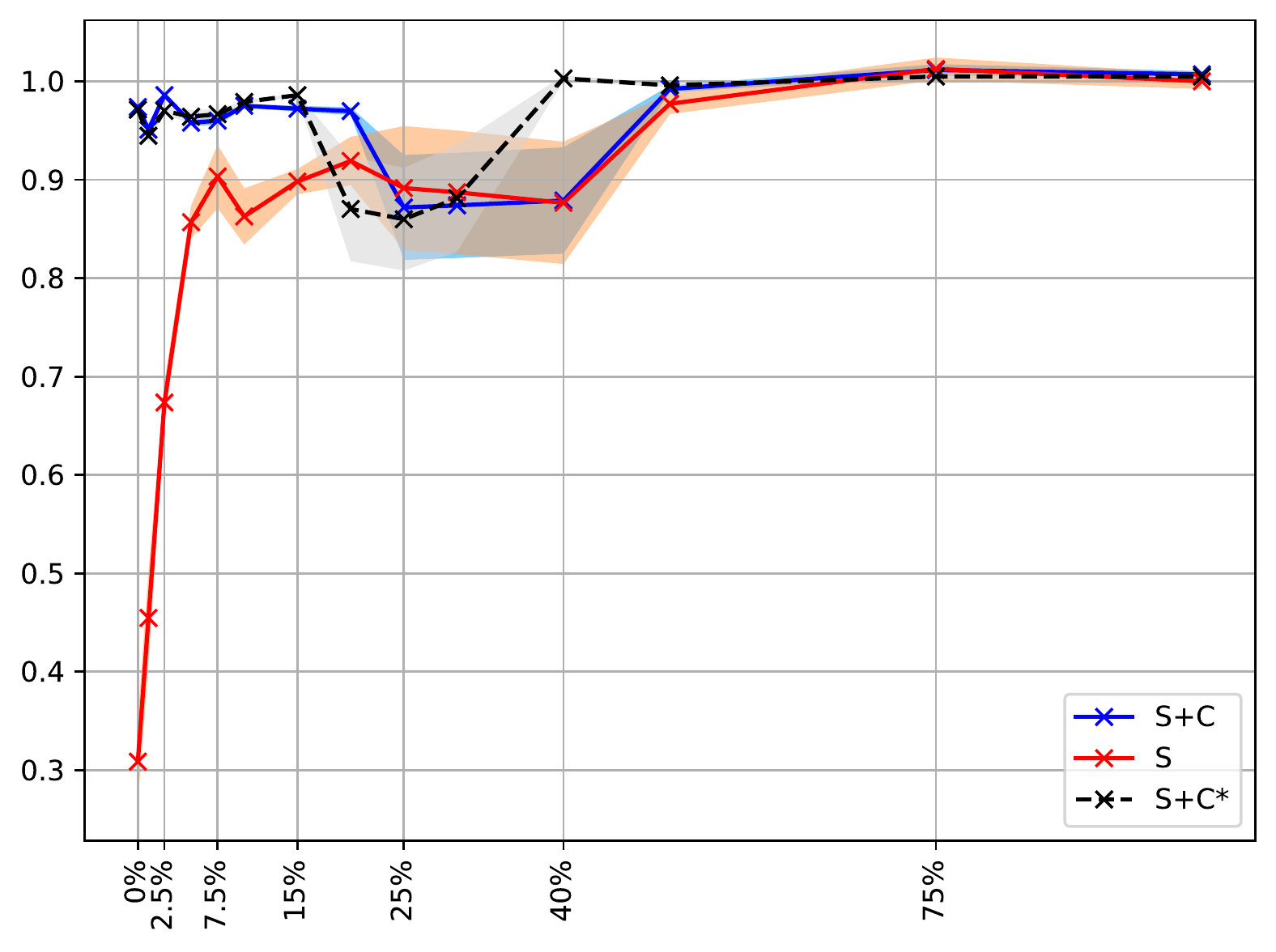}
         \caption{F1 (macro), DigestPath 2019}
         \label{fig:c_s_f1mac_digestpath2019}
     \end{subfigure}
     \begin{subfigure}[b]{0.4\textwidth}
         \centering
         \includegraphics[width=\textwidth]{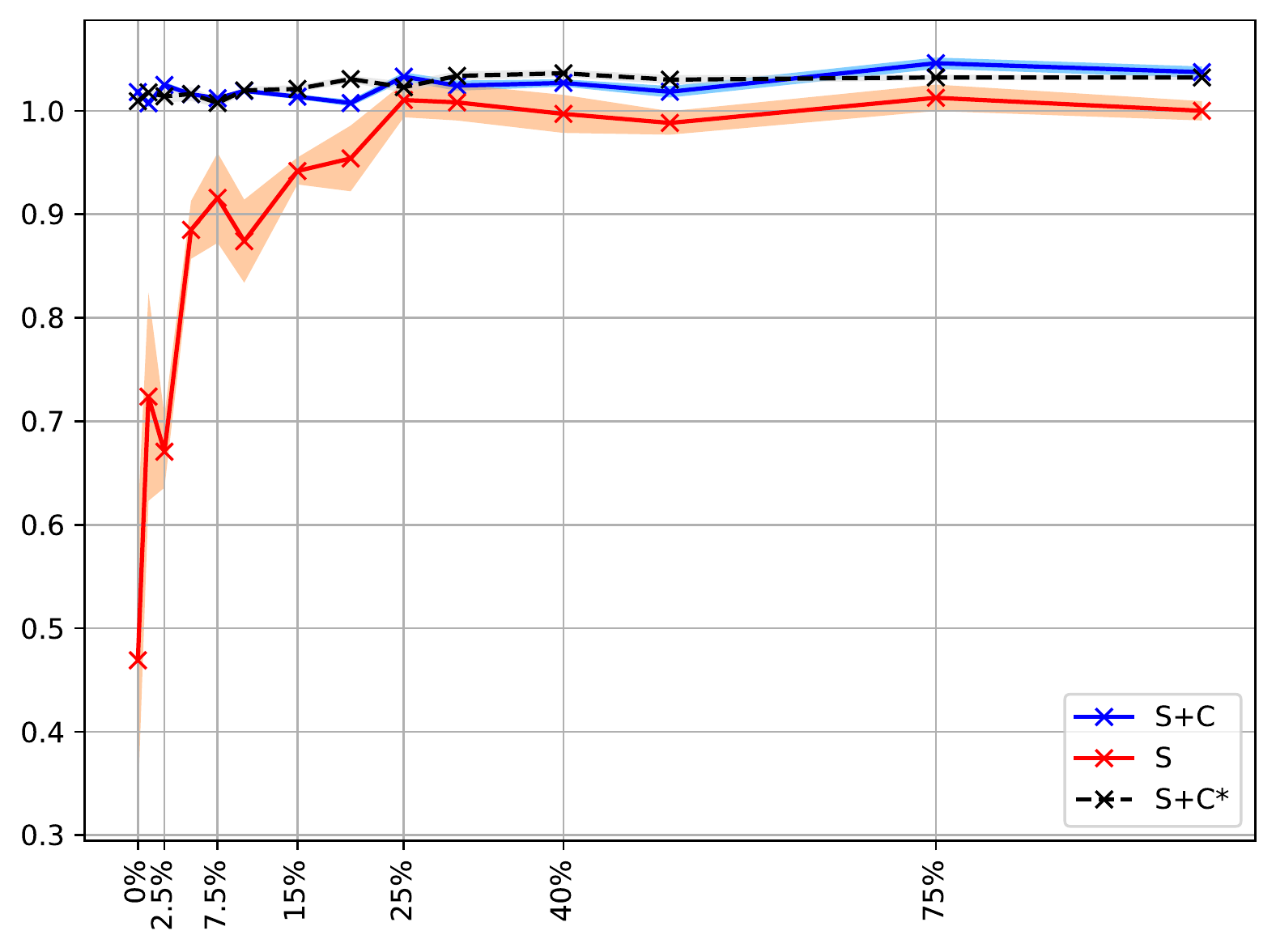}
         \caption{F1 (micro), DigestPath 2019}
         \label{fig:c_s_f1mic_digestpath2019}
     \end{subfigure}

        \caption{Comparison of training performance between using only segmentation (S) patches, both segmentation and classification (\textbf{S+C}) images, and varying the amount of segmentation patches while using the complete set of classification patches (\textbf{S+C*}). These results are normalized to s=100\%. For the raw results, please refer to \ref{tab:classification_head_results_raw}. For the ICIAR BACH 2018 dataset, the following number of images  correspond to each percentage setting, respectively (0\% to 100\%): 0, 6, 15, 30, 46, 61, 92, 123, 153, 184, 246, 307, 461, and 615 segmentation-level, and 400, 396, 390, 380, 370, 360, 340, 320, 300, 280, 240, 200, 100, 0 patch-level images. For the Gleason2019 dataset, the following number of images  correspond to each percentage setting, respectively (0\% to 100\%): 0, 12, 31, 63, 95, 127, 191, 254, 318, 382, 509, 637, 955, 1274 segmentation-level, and 1774, 1757, 1730, 1686, 1641, 1597, 1508, 1420, 1331, 1242, 1065, 887,  444, 0 patch-level images. For the DigestPath2019 dataset, the following number of images  correspond to each percentage setting, respectively (0\% to 100\%): 0, 16, 40, 81, 122, 163, 244, 326, 407, 489, 652, 815, 1222, 1630 segmentation-level, and 1764, 1747, 1720, 1676, 1632, 1588, 1500, 1412, 1323, 1235, 1059, 882, 441, 0 patch-level images.  }
        \label{fig:c_s_comparison}
\end{figure*}

We summarize our segmentation results in Fig. \ref{fig:c_s_comparison}. The $\times$ and $\bullet$ markers in Fig. \ref{fig:c_s_comparison} are mean values obtained from 5 experiments, and the shaded regions represent the standard deviations. The horizontal axis indicates the $s$, whereas the vertical axis is the relative performance with respect to the performance at $s=100$, the case where we only use segmentation patches. We chose this presentation as the absolute values are not as significant as difference in performance as we use more segmentation patches. For the raw metric results, please refer to the Table \ref{tab:segmentation_results_raw}.

We achieve a large improvement in performance when the percentage of segmentation level patches are low compared to the patch level images, and performances of the two methods converge as we increase the number of segmentation patches. This is desirable since we are able to achieve similar performance with data acquired much more cheaply. More specifically, there is a significant performance gap ($\geq 15\%$ for both F1 metrics) between our method (\textbf{S+C} or \textbf{S+C*}) and the \textbf{S} setting when we use $\leq 10\%$ of segmentation patches. In addition, if we keep the classification patches while increasing the amount of segmentation patches (\textbf{S+C*}), we still observe gains, indicating that the method can work either in low or high data settings. We also observe less variation in performance with our method, since the use of more data acts as a regularizer. This can also be visually observed in Fig. \ref{fig:visual_samples_bach}, where the predictions are more stable as we force a more general feature representation with image level data that prevents radical changes in decisions when the network is trained with more data. For instance, predictions of the \textbf{S} setting exhibit large variability in close spatial proximity, such as the blue to green class change between neighboring regions that visually appear similar on the whole slide image. In contrast, in \textbf{S+C} setting such variability is less apparent, as the trained network is able to extract semantic information more robustly to prevent inconsistent decisions for visually and contextually similar regions.

We present output segmentation masks for settings with $s\leq10$, and compare \textbf{S}, \textbf{S+C} and \textbf{S+C*} in figures  \ref{fig:visual_samples_bach}, \ref{fig:visual_samples_gleason2019}, and \ref{fig:visual_samples_digestpath2019}. For the ICIAR BACH 2018 dataset, we use the color mapping provided by the challenge organizers \citep{aresta2019bach}, where white is normal (background) tissue, red is benign, green is in situ, and the blue is the invasive class. For Gleason2019 and DigestPath2019, we use the jet colormap, which maps integers (0 and 1 for Gleason2019 and 0 to 4 for DigestPath2019) into their respective RGB counterparts. We omit sample outputs for settings with $s>10$ even though our method remains comparable or better to \textbf{S}, since beyond a threshold, performance improvements remain marginal when the additional time spent for annotating a segmentation vs. classification patch is considered.  Whereas the \textbf{S} setting only is able to achieve reasonable performance at $s=10$, \textbf{S+C} consistently performs well under $s<10$. \textbf{S+C} is also less prone to make mistakes, such as labeling two very close regions that look visually similar as different classes, which is an indication of noisy feature learning that suffers from overfitting.  

For reference, we compare our method's results with the state-of-the-art for all datasets presented in this text. \cite{ciga2019multi} achieve a challenge-specific score of 68\% on the ICIAR BACH 2018, whereas our method obtains a score of 54\%. \cite{9098678} achieve 67.9\% Dice (F1) score with a U-Net on DigestPath2019, whereas we only achieve 42\%. Finally, \cite{zhang2020gleason} obtain a 75\% Dice to our 41\%. In addition to using only half the dataset for training and downsampling each image to $128\times128$ pixels, which amounts to $8\times$ downsampling per extracted patch, we argue that the large performance gap is due to  over-engineering to the specific dataset. For instance, \cite{ciga2019multi} use a specialized domain adaptation regularization technique coupled with heavy data augmentations using a network with 116 million trainable parameters to improve their baseline result of 42\%. Similarly, \cite{9098678} use a customized network to improve their baseline results by more than 10\%. In contrast, we do not use any customized network, use a smaller network with 8 million trainable parameters, ignore some segmentation mask ground truth to generate classification patches, and do not perform any data-specific fine-tuning. The results obtained for s =100\%, which is equivalent to a standard U-net model trained on all available segmentation patches, provide a more useful baseline and we believe our  method  can  be  incorporated  into  more  sophisticated  net-works for furthering their performance. We believe our method can be incorporated into more sophisticated networks for furthering their performance. Finally, while F1 scores are accepted universally and can be used for direct comparison to other methods, they may be insufficient in representing the nuances between different methods. For instance, class-wise F1 scores, as well as the intersection over union, improve as we add more segmentation patches, which indicate the utility of having a combination of these two data sources. Further experimental results can be accessed from the supplementary material of this text.

Combining segmentation and classification data is aimed at reducing the inter-expert disagreement and minimizing the effect of errors made during the pixel-level annotation step. However, a digital pathology image is a 2D projection of a 3D tissue volume  and obtaining the ``perfect" ground truth is, therefore, not possible. This ambiguity will be present and cannot be mitigated by our method regardless of the quality (e.g., resolution) of images, time allocated for annotation, number of classification patches, or the annotator's experience.
%3d vs 2d: ground truth is a relative term. there's no such thing as a perfect ground truth. even if we improve magnification, labeler takes infinite time still not 100%. even when annotator goes 100% around image it's not 100% correct because partial volume. }

Finally, the proposed method is not capable of correcting errors if the incorrect annotations are of the same class as the image level label for the given patch, or if there are multiple incorrect annotations on a predicted patch. In our experiments, we observed that the probability of making this type of error decreases as the training set size increases. In addition, we observed that the data collection process is crucial to the success of the pipeline. The model is able to distinguish between the background and the foreground given only a few segmentation patches, by learning low-level features (e.g., edge or blob extractors) to outline the boundaries of the foreground structure, however, the correct class assignments of the segmented regions are more challenging compared to the background and foreground separation. Since the error signal from the classification layer does not specify which pixels should be corrected, image level data should be collected where the majority of the pixels on the patch are either background or the foreground class assigned to that patch.

\section{Conclusions}

%remove classification graphs (only keep cfs mtx)

%ethical: all open access datasets..reb approval was not needed, each dataset was individually clear by their respective rebs.. to reviewer: we assume by ethical you mean REBs.. (put reb links)

%3d vs 2d: ground truth is a relative term. there's no such thing as a perfect ground truth. even if we improve magnification, labeler takes infinite time still not 100%. even when annotator goes 100% around image it's not 100% correct because partial volume. 

%Fiji or QuPath?: don't put anything on the paper, incorporating is complicated. sedeen long term incorporation will be done. since on github, anybody can do it we don't have resources. as a group we are looking at how to do it package it..

%lay summary: grade 3 reading ability. this will reduce the effort required for labeling. this allows dig path imgs annot more efficiently, increasing amount of label in these datasets. quicker, cheaper, easier. to reviewer: we think relatively easy to understand but added these 2-3 sentences..  

%figure caption: make figure bifurcation/path going through much clearer. bunch of images...

%https://iciar2018-challenge.grand-challenge.org/evaluation/2e97cc68-95fe-45f6-97c6-6f48297c0e03/

In this paper, we presented a method that can expedite medical image annotation by labeling a few segmentation level patches, and the majority of the training data is composed of easier-to-label image level patches. With the currently available architectures, training with two types of patches is not possible, hence we hope that our work can help researchers in expediting medical imaging tasks, and can be used as a baseline for improving techniques that amalgamate different types of training data. We validated the efficacy of our method in settings where we have a large imbalance between segmentation and image level patches. Our method can be used to expedite tasks at the data acquisition stage, or it can be used for utilizing previously acquired data that only includes image level patches for segmentation tasks by drawing boundaries for a few samples from each class in the dataset such as BreakHis cancer classification task \citep{spanhol2015dataset}. Finally, we acknowledge the shortcomings of our method, namely failure to separate the feature representation of background and the foreground when training patches include very small regions of interest, and the inability to train for large patches (e.g., $512\times512$ pixels) that is critical in breast histopathology, and aim to address these issues in future work. 

\section*{Conflict of interest} \begin{flushleft} We have no conflict of interest to declare. \end{flushleft}

\section*{Acknowledgments} \begin{flushleft} This work was funded by Canadian Cancer Society (grant \#705772) and NSERC RGPIN-2016-06283. We thank our collaborating pathologists Dina Bassiouny and Sharon Nofech-Mozes for the useful discussions on time required for annotating WSIs.\end{flushleft}

%%Harvard
\bibliographystyle{model2-names.bst}\biboptions{authoryear}
\bibliography{refs}

\begin{thebibliography}{32}
\expandafter\ifx\csname natexlab\endcsname\relax\def\natexlab#1{#1}\fi
\providecommand{\url}[1]{\texttt{#1}}
\providecommand{\href}[2]{#2}
\providecommand{\path}[1]{#1}
\providecommand{\DOIprefix}{doi:}
\providecommand{\ArXivprefix}{arXiv:}
\providecommand{\URLprefix}{URL: }
\providecommand{\Pubmedprefix}{pmid:}
\providecommand{\doi}[1]{\href{http://dx.doi.org/#1}{\path{#1}}}
\providecommand{\Pubmed}[1]{\href{pmid:#1}{\path{#1}}}
\providecommand{\bibinfo}[2]{#2}
\ifx\xfnm\relax \def\xfnm[#1]{\unskip,\space#1}\fi
%Type = Article
\bibitem[{Ahn et~al.(2019)Ahn, Kumar, Fulham, Feng and Kim}]{AHN2019140}
\bibinfo{author}{Ahn, E.}, \bibinfo{author}{Kumar, A.},
  \bibinfo{author}{Fulham, M.}, \bibinfo{author}{Feng, D.},
  \bibinfo{author}{Kim, J.}, \bibinfo{year}{2019}.
\newblock \bibinfo{title}{Convolutional sparse kernel network for unsupervised
  medical image analysis}.
\newblock \bibinfo{journal}{Medical Image Analysis} \bibinfo{volume}{56},
  \bibinfo{pages}{140 -- 151}.
\newblock \URLprefix
  \url{http://www.sciencedirect.com/science/article/pii/S1361841518306868},
  \DOIprefix\doi{https://doi.org/10.1016/j.media.2019.06.005}.
%Type = Article
\bibitem[{Aresta et~al.(2019)Aresta, Ara{\'u}jo, Kwok, Chennamsetty, Safwan,
  Alex, Marami, Prastawa, Chan, Donovan et~al.}]{aresta2019bach}
\bibinfo{author}{Aresta, G.}, \bibinfo{author}{Ara{\'u}jo, T.},
  \bibinfo{author}{Kwok, S.}, \bibinfo{author}{Chennamsetty, S.S.},
  \bibinfo{author}{Safwan, M.}, \bibinfo{author}{Alex, V.},
  \bibinfo{author}{Marami, B.}, \bibinfo{author}{Prastawa, M.},
  \bibinfo{author}{Chan, M.}, \bibinfo{author}{Donovan, M.}, et~al.,
  \bibinfo{year}{2019}.
\newblock \bibinfo{title}{Bach: Grand challenge on breast cancer histology
  images}.
\newblock \bibinfo{journal}{Medical image analysis} .
%Type = Inproceedings
\bibitem[{Bokhorst et~al.(2019)Bokhorst, Pinckaers, {van Zwam}, Nagtegaal, {van
  der Laak} and Ciompi}]{Bokhorst19}
\bibinfo{author}{Bokhorst, J.}, \bibinfo{author}{Pinckaers, H.},
  \bibinfo{author}{{van Zwam}, P.}, \bibinfo{author}{Nagtegaal, I.},
  \bibinfo{author}{{van der Laak}, J.}, \bibinfo{author}{Ciompi, F.},
  \bibinfo{year}{2019}.
\newblock \bibinfo{title}{Learning from sparsely annotated data for semantic
  segmentation in histopathology images}, in: \bibinfo{booktitle}{Proceedings
  of the 2nd International Conference on Medical Imaging with Deep Learning},
  pp. \bibinfo{pages}{84--91}.
%Type = Article
\bibitem[{Chen et~al.(2019)Chen, Bentley, Mori, Misawa, Fujiwara and
  Rueckert}]{CHEN2019101539}
\bibinfo{author}{Chen, L.}, \bibinfo{author}{Bentley, P.},
  \bibinfo{author}{Mori, K.}, \bibinfo{author}{Misawa, K.},
  \bibinfo{author}{Fujiwara, M.}, \bibinfo{author}{Rueckert, D.},
  \bibinfo{year}{2019}.
\newblock \bibinfo{title}{Self-supervised learning for medical image analysis
  using image context restoration}.
\newblock \bibinfo{journal}{Medical Image Analysis} \bibinfo{volume}{58},
  \bibinfo{pages}{101539}.
\newblock \URLprefix
  \url{http://www.sciencedirect.com/science/article/pii/S1361841518304699},
  \DOIprefix\doi{https://doi.org/10.1016/j.media.2019.101539}.
%Type = Incollection
\bibitem[{Ciga et~al.(2019)Ciga, Chen and Martel}]{ciga2019multi}
\bibinfo{author}{Ciga, O.}, \bibinfo{author}{Chen, J.},
  \bibinfo{author}{Martel, A.}, \bibinfo{year}{2019}.
\newblock \bibinfo{title}{Multi-layer domain adaptation for deep convolutional
  networks}, in: \bibinfo{booktitle}{Domain Adaptation and Representation
  Transfer and Medical Image Learning with Less Labels and Imperfect Data}.
  \bibinfo{publisher}{Springer}, pp. \bibinfo{pages}{20--27}.
%Type = Inproceedings
\bibitem[{Crowston(2012)}]{10.1007/978-3-642-35142-6_14}
\bibinfo{author}{Crowston, K.}, \bibinfo{year}{2012}.
\newblock \bibinfo{title}{Amazon mechanical turk: A research tool for
  organizations and information systems scholars}, in:
  \bibinfo{editor}{Bhattacherjee, A.}, \bibinfo{editor}{Fitzgerald, B.} (Eds.),
  \bibinfo{booktitle}{Shaping the Future of ICT Research. Methods and
  Approaches}, \bibinfo{publisher}{Springer Berlin Heidelberg},
  \bibinfo{address}{Berlin, Heidelberg}. pp. \bibinfo{pages}{210--221}.
%Type = Inproceedings
\bibitem[{Deng et~al.(2009)Deng, Dong, Socher, Li, Li and
  Fei-Fei}]{imagenet_cvpr09}
\bibinfo{author}{Deng, J.}, \bibinfo{author}{Dong, W.},
  \bibinfo{author}{Socher, R.}, \bibinfo{author}{Li, L.J.},
  \bibinfo{author}{Li, K.}, \bibinfo{author}{Fei-Fei, L.},
  \bibinfo{year}{2009}.
\newblock \bibinfo{title}{{ImageNet: A Large-Scale Hierarchical Image
  Database}}, in: \bibinfo{booktitle}{CVPR09}.
%Type = Article
\bibitem[{Gidaris et~al.(2018)Gidaris, Singh and
  Komodakis}]{gidaris2018unsupervised}
\bibinfo{author}{Gidaris, S.}, \bibinfo{author}{Singh, P.},
  \bibinfo{author}{Komodakis, N.}, \bibinfo{year}{2018}.
\newblock \bibinfo{title}{Unsupervised representation learning by predicting
  image rotations}.
\newblock \bibinfo{journal}{arXiv preprint arXiv:1803.07728} .
%Type = Inproceedings
\bibitem[{He et~al.(2016)He, Zhang, Ren and Sun}]{he2016deep}
\bibinfo{author}{He, K.}, \bibinfo{author}{Zhang, X.}, \bibinfo{author}{Ren,
  S.}, \bibinfo{author}{Sun, J.}, \bibinfo{year}{2016}.
\newblock \bibinfo{title}{Deep residual learning for image recognition}, in:
  \bibinfo{booktitle}{Proceedings of the IEEE conference on computer vision and
  pattern recognition}, pp. \bibinfo{pages}{770--778}.
%Type = Article
\bibitem[{Hou et~al.(2019)Hou, Nguyen, Kanevsky, Samaras, Kurc, Zhao, Gupta,
  Gao, Chen, Foran et~al.}]{hou2019sparse}
\bibinfo{author}{Hou, L.}, \bibinfo{author}{Nguyen, V.},
  \bibinfo{author}{Kanevsky, A.B.}, \bibinfo{author}{Samaras, D.},
  \bibinfo{author}{Kurc, T.M.}, \bibinfo{author}{Zhao, T.},
  \bibinfo{author}{Gupta, R.R.}, \bibinfo{author}{Gao, Y.},
  \bibinfo{author}{Chen, W.}, \bibinfo{author}{Foran, D.}, et~al.,
  \bibinfo{year}{2019}.
\newblock \bibinfo{title}{Sparse autoencoder for unsupervised nucleus detection
  and representation in histopathology images}.
\newblock \bibinfo{journal}{Pattern Recognition} \bibinfo{volume}{86},
  \bibinfo{pages}{188--200}.
%Type = Article
\bibitem[{Hu et~al.(2018)Hu, Tang, Eric, Chang, Fan, Lai and
  Xu}]{hu2018unsupervised}
\bibinfo{author}{Hu, B.}, \bibinfo{author}{Tang, Y.}, \bibinfo{author}{Eric,
  I.}, \bibinfo{author}{Chang, C.}, \bibinfo{author}{Fan, Y.},
  \bibinfo{author}{Lai, M.}, \bibinfo{author}{Xu, Y.}, \bibinfo{year}{2018}.
\newblock \bibinfo{title}{Unsupervised learning for cell-level visual
  representation in histopathology images with generative adversarial
  networks}.
\newblock \bibinfo{journal}{IEEE Journal of Biomedical and Health Informatics}
  \bibinfo{volume}{23}, \bibinfo{pages}{1316--1328}.
%Type = Inproceedings
\bibitem[{Ioffe and Szegedy(2015)}]{ioffe2015batch}
\bibinfo{author}{Ioffe, S.}, \bibinfo{author}{Szegedy, C.},
  \bibinfo{year}{2015}.
\newblock \bibinfo{title}{Batch normalization: Accelerating deep network
  training by reducing internal covariate shift}, pp.
  \bibinfo{pages}{448--456}.
\newblock \URLprefix \url{http://jmlr.org/proceedings/papers/v37/ioffe15.pdf}.
%Type = Article
\bibitem[{Li et~al.(2019a)Li, Wang, Liu, Latecki, Wang and
  Huang}]{Li2019weakly}
\bibinfo{author}{Li, C.}, \bibinfo{author}{Wang, X.}, \bibinfo{author}{Liu,
  W.}, \bibinfo{author}{Latecki, L.J.}, \bibinfo{author}{Wang, B.},
  \bibinfo{author}{Huang, J.}, \bibinfo{year}{2019}a.
\newblock \bibinfo{title}{Weakly supervised mitosis detection in breast
  histopathology images using concentric loss}.
\newblock \bibinfo{journal}{Medical image analysis} \bibinfo{volume}{53},
  \bibinfo{pages}{165--178}.
%Type = Inproceedings
\bibitem[{Li et~al.(2019b)Li, Yang, Huang, Da, Yang, Hu, Duan, Wang and
  Li}]{li2019signet}
\bibinfo{author}{Li, J.}, \bibinfo{author}{Yang, S.}, \bibinfo{author}{Huang,
  X.}, \bibinfo{author}{Da, Q.}, \bibinfo{author}{Yang, X.},
  \bibinfo{author}{Hu, Z.}, \bibinfo{author}{Duan, Q.}, \bibinfo{author}{Wang,
  C.}, \bibinfo{author}{Li, H.}, \bibinfo{year}{2019}b.
\newblock \bibinfo{title}{Signet ring cell detection with a semi-supervised
  learning framework}, in: \bibinfo{booktitle}{International Conference on
  Information Processing in Medical Imaging}, \bibinfo{organization}{Springer}.
  pp. \bibinfo{pages}{842--854}.
%Type = Inproceedings
\bibitem[{{Li} et~al.(2020){Li}, {Xu}, {Wang}, {Zhou} and {Zhang}}]{9098678}
\bibinfo{author}{{Li}, Y.}, \bibinfo{author}{{Xu}, Z.},
  \bibinfo{author}{{Wang}, Y.}, \bibinfo{author}{{Zhou}, H.},
  \bibinfo{author}{{Zhang}, Q.}, \bibinfo{year}{2020}.
\newblock \bibinfo{title}{Su-net and du-net fusion for tumour segmentation in
  histopathology images}, in: \bibinfo{booktitle}{2020 IEEE 17th International
  Symposium on Biomedical Imaging (ISBI)}, pp. \bibinfo{pages}{461--465}.
%Type = Inproceedings
\bibitem[{Macenko et~al.(2009)Macenko, Niethammer, Marron, Borland, Woosley,
  Guan, Schmitt and Thomas}]{macenko2009method}
\bibinfo{author}{Macenko, M.}, \bibinfo{author}{Niethammer, M.},
  \bibinfo{author}{Marron, J.S.}, \bibinfo{author}{Borland, D.},
  \bibinfo{author}{Woosley, J.T.}, \bibinfo{author}{Guan, X.},
  \bibinfo{author}{Schmitt, C.}, \bibinfo{author}{Thomas, N.E.},
  \bibinfo{year}{2009}.
\newblock \bibinfo{title}{A method for normalizing histology slides for
  quantitative analysis}, in: \bibinfo{booktitle}{2009 IEEE International
  Symposium on Biomedical Imaging: From Nano to Macro},
  \bibinfo{organization}{IEEE}. pp. \bibinfo{pages}{1107--1110}.
%Type = Inbook
\bibitem[{Mehta et~al.(2018)Mehta, Mercan, Bartlett, Weaver, Elmore and
  Shapiro}]{mehta2018ynet}
\bibinfo{author}{Mehta, S.}, \bibinfo{author}{Mercan, E.},
  \bibinfo{author}{Bartlett, J.}, \bibinfo{author}{Weaver, D.},
  \bibinfo{author}{Elmore, J.}, \bibinfo{author}{Shapiro, L.},
  \bibinfo{year}{2018}.
\newblock \bibinfo{title}{Y-Net: Joint Segmentation and Classification for
  Diagnosis of Breast Biopsy Images: 21st International Conference, Granada,
  Spain, September 16–20, 2018, Proceedings, Part II}.
\newblock pp. \bibinfo{pages}{893--901}.
\newblock \DOIprefix\doi{10.1007/978-3-030-00934-2_99}.
%Type = Inproceedings
\bibitem[{Noroozi and Favaro(2016)}]{noroozi2016unsupervised}
\bibinfo{author}{Noroozi, M.}, \bibinfo{author}{Favaro, P.},
  \bibinfo{year}{2016}.
\newblock \bibinfo{title}{Unsupervised learning of visual representations by
  solving jigsaw puzzles}, in: \bibinfo{booktitle}{European Conference on
  Computer Vision}, \bibinfo{organization}{Springer}. pp.
  \bibinfo{pages}{69--84}.
%Type = Article
\bibitem[{{\O}rting et~al.(2019){\O}rting, Doyle, Hirth, van Hilten, Inel,
  Madan, Mavridis, Spiers and Cheplygina}]{rting2019ASO}
\bibinfo{author}{{\O}rting, S.N.}, \bibinfo{author}{Doyle, A.J.},
  \bibinfo{author}{Hirth, M.}, \bibinfo{author}{van Hilten, A.},
  \bibinfo{author}{Inel, O.}, \bibinfo{author}{Madan, C.R.},
  \bibinfo{author}{Mavridis, P.}, \bibinfo{author}{Spiers, H.},
  \bibinfo{author}{Cheplygina, V.}, \bibinfo{year}{2019}.
\newblock \bibinfo{title}{A survey of crowdsourcing in medical image analysis}.
\newblock \bibinfo{journal}{ArXiv} \bibinfo{volume}{abs/1902.09159}.
%Type = Inproceedings
\bibitem[{Qu et~al.(2019)Qu, Wu, Huang, Yi, Riedlinger, De and
  Metaxas}]{Qu2019weakly}
\bibinfo{author}{Qu, H.}, \bibinfo{author}{Wu, P.}, \bibinfo{author}{Huang,
  Q.}, \bibinfo{author}{Yi, J.}, \bibinfo{author}{Riedlinger, G.M.},
  \bibinfo{author}{De, S.}, \bibinfo{author}{Metaxas, D.N.},
  \bibinfo{year}{2019}.
\newblock \bibinfo{title}{Weakly supervised deep nuclei segmentation using
  points annotation in histopathology images}, in:
  \bibinfo{booktitle}{International Conference on Medical Imaging with Deep
  Learning}, pp. \bibinfo{pages}{390--400}.
%Type = Article
\bibitem[{Rajchl et~al.(2017)Rajchl, Koch, Ledig, Passerat-Palmbach, Misawa,
  Mori and Rueckert}]{rajchl2017employing}
\bibinfo{author}{Rajchl, M.}, \bibinfo{author}{Koch, L.M.},
  \bibinfo{author}{Ledig, C.}, \bibinfo{author}{Passerat-Palmbach, J.},
  \bibinfo{author}{Misawa, K.}, \bibinfo{author}{Mori, K.},
  \bibinfo{author}{Rueckert, D.}, \bibinfo{year}{2017}.
\newblock \bibinfo{title}{Employing weak annotations for medical image analysis
  problems}.
\newblock \bibinfo{journal}{arXiv preprint arXiv:1708.06297} .
%Type = Inproceedings
\bibitem[{Remez et~al.(2018)Remez, Huang and Brown}]{remez2018learning}
\bibinfo{author}{Remez, T.}, \bibinfo{author}{Huang, J.},
  \bibinfo{author}{Brown, M.}, \bibinfo{year}{2018}.
\newblock \bibinfo{title}{Learning to segment via cut-and-paste}, in:
  \bibinfo{booktitle}{Proceedings of the European Conference on Computer Vision
  (ECCV)}, pp. \bibinfo{pages}{37--52}.
%Type = Inproceedings
\bibitem[{Ronneberger et~al.(2015)Ronneberger, Fischer and
  Brox}]{10.1007/978-3-319-24574-4_28}
\bibinfo{author}{Ronneberger, O.}, \bibinfo{author}{Fischer, P.},
  \bibinfo{author}{Brox, T.}, \bibinfo{year}{2015}.
\newblock \bibinfo{title}{U-net: Convolutional networks for biomedical image
  segmentation}, in: \bibinfo{editor}{Navab, N.}, \bibinfo{editor}{Hornegger,
  J.}, \bibinfo{editor}{Wells, W.M.}, \bibinfo{editor}{Frangi, A.F.} (Eds.),
  \bibinfo{booktitle}{Medical Image Computing and Computer-Assisted
  Intervention -- MICCAI 2015}, \bibinfo{publisher}{Springer International
  Publishing}, \bibinfo{address}{Cham}. pp. \bibinfo{pages}{234--241}.
%Type = Inproceedings
\bibitem[{Seth et~al.(2019)Seth, Akbar, Nofech-Mozes, Salama and
  Martel}]{seth2018automated}
\bibinfo{author}{Seth, N.}, \bibinfo{author}{Akbar, S.},
  \bibinfo{author}{Nofech-Mozes, S.}, \bibinfo{author}{Salama, S.},
  \bibinfo{author}{Martel, A.L.}, \bibinfo{year}{2019}.
\newblock \bibinfo{title}{Automated segmentation of {DCIS} in whole slide
  images}, in: \bibinfo{booktitle}{European Congress on Digital Pathology ECDP
  2019}, pp. \bibinfo{pages}{67--74}.
%Type = Article
\bibitem[{Spanhol et~al.(2015)Spanhol, Oliveira, Petitjean and
  Heutte}]{spanhol2015dataset}
\bibinfo{author}{Spanhol, F.A.}, \bibinfo{author}{Oliveira, L.S.},
  \bibinfo{author}{Petitjean, C.}, \bibinfo{author}{Heutte, L.},
  \bibinfo{year}{2015}.
\newblock \bibinfo{title}{A dataset for breast cancer histopathological image
  classification}.
\newblock \bibinfo{journal}{IEEE Transactions on Biomedical Engineering}
  \bibinfo{volume}{63}, \bibinfo{pages}{1455--1462}.
%Type = Article
\bibitem[{Srinidhi et~al.(2019)Srinidhi, Ciga and Martel}]{srinidhi2019deep}
\bibinfo{author}{Srinidhi, C.L.}, \bibinfo{author}{Ciga, O.},
  \bibinfo{author}{Martel, A.L.}, \bibinfo{year}{2019}.
\newblock \bibinfo{title}{Deep neural network models for computational
  histopathology: A survey}.
\newblock \bibinfo{journal}{arXiv preprint arXiv:1912.12378} .
%Type = Inproceedings
\bibitem[{Tajbakhsh et~al.(2019)Tajbakhsh, Hu, Cao, Yan, Xiao, Lu, Liang,
  Terzopoulos and Ding}]{5b38f8cb124741559716f8dc05e36932}
\bibinfo{author}{Tajbakhsh, N.}, \bibinfo{author}{Hu, Y.},
  \bibinfo{author}{Cao, J.}, \bibinfo{author}{Yan, X.}, \bibinfo{author}{Xiao,
  Y.}, \bibinfo{author}{Lu, Y.}, \bibinfo{author}{Liang, J.},
  \bibinfo{author}{Terzopoulos, D.}, \bibinfo{author}{Ding, X.},
  \bibinfo{year}{2019}.
\newblock \bibinfo{title}{Surrogate supervision for medical image analysis:
  Effective deep learning from limited quantities of labeled data}, in:
  \bibinfo{booktitle}{ISBI 2019 - 2019 IEEE International Symposium on
  Biomedical Imaging}, \bibinfo{publisher}{IEEE Computer Society}. pp.
  \bibinfo{pages}{1251--1255}.
\newblock \DOIprefix\doi{10.1109/ISBI.2019.8759553}.
%Type = Misc
\bibitem[{Taleb et~al.(2019)Taleb, Lippert, Klein and
  Nabi}]{taleb2019multimodal}
\bibinfo{author}{Taleb, A.}, \bibinfo{author}{Lippert, C.},
  \bibinfo{author}{Klein, T.}, \bibinfo{author}{Nabi, M.},
  \bibinfo{year}{2019}.
\newblock \bibinfo{title}{Multimodal self-supervised learning for medical image
  analysis}.
\newblock \href{http://arxiv.org/abs/1912.05396}{\tt arXiv:1912.05396}.
%Type = Article
\bibitem[{Wong et~al.(2018)Wong, Syeda-Mahmood and Moradi}]{WONG2018105}
\bibinfo{author}{Wong, K.C.}, \bibinfo{author}{Syeda-Mahmood, T.},
  \bibinfo{author}{Moradi, M.}, \bibinfo{year}{2018}.
\newblock \bibinfo{title}{Building medical image classifiers with very limited
  data using segmentation networks}.
\newblock \bibinfo{journal}{Medical Image Analysis} \bibinfo{volume}{49},
  \bibinfo{pages}{105 -- 116}.
\newblock \URLprefix
  \url{http://www.sciencedirect.com/science/article/pii/S1361841518305516},
  \DOIprefix\doi{https://doi.org/10.1016/j.media.2018.07.010}.
%Type = Article
\bibitem[{Xu et~al.(2015)Xu, Xiang, Liu, Gilmore, Wu, Tang and
  Madabhushi}]{xu2015stacked}
\bibinfo{author}{Xu, J.}, \bibinfo{author}{Xiang, L.}, \bibinfo{author}{Liu,
  Q.}, \bibinfo{author}{Gilmore, H.}, \bibinfo{author}{Wu, J.},
  \bibinfo{author}{Tang, J.}, \bibinfo{author}{Madabhushi, A.},
  \bibinfo{year}{2015}.
\newblock \bibinfo{title}{Stacked sparse autoencoder (ssae) for nuclei
  detection on breast cancer histopathology images}.
\newblock \bibinfo{journal}{IEEE Transactions on Medical Imaging}
  \bibinfo{volume}{35}, \bibinfo{pages}{119--130}.
%Type = Article
\bibitem[{Yang et~al.(2018)Yang, Zhang, Zhao, Zheng, Liang, Ying, Ahuja and
  Chen}]{Yang2018boxnet}
\bibinfo{author}{Yang, L.}, \bibinfo{author}{Zhang, Y.}, \bibinfo{author}{Zhao,
  Z.}, \bibinfo{author}{Zheng, H.}, \bibinfo{author}{Liang, P.},
  \bibinfo{author}{Ying, M.T.}, \bibinfo{author}{Ahuja, A.T.},
  \bibinfo{author}{Chen, D.Z.}, \bibinfo{year}{2018}.
\newblock \bibinfo{title}{Boxnet: Deep learning based biomedical image
  segmentation using boxes only annotation}.
\newblock \bibinfo{journal}{arXiv preprint arXiv:1806.00593} .
%Type = Article
\bibitem[{Zhang et~al.(2020)Zhang, Zhang, Song, Shen and
  Yang}]{zhang2020gleason}
\bibinfo{author}{Zhang, Y.h.}, \bibinfo{author}{Zhang, J.},
  \bibinfo{author}{Song, Y.}, \bibinfo{author}{Shen, C.},
  \bibinfo{author}{Yang, G.}, \bibinfo{year}{2020}.
\newblock \bibinfo{title}{Gleason score prediction using deep learning in
  tissue microarray image}.
\newblock \bibinfo{journal}{arXiv preprint arXiv:2005.04886} .

\end{thebibliography}

\clearpage

\onecolumn
\newgeometry{left=0.5cm, right=0.5cm, top=0.5cm}

\appendix

\section{Architectures}

We use the third layer output (out of a total of four layers) of the Resnet-18 architecture (pretrained on ILSVRC dataset) for encoding input images due to the small input size ($128\times128$ pixels) we use in our experiments, as opposed to the default input size for Resnet ($224\times224$ pixels). The classification and segmentation modules are given below, where we replace the fully connected linear layers with $1\times1$ convolutions. 

\begin{table}[h]
\centering
   \begin{tabular}{c}
   Decoder \\
   \toprule \midrule
      Input $256 \times 8 \times 8$ matrix \\
      \midrule
      Conv2d $3 \times 3$, stride 1, padding 1, $256 > 64$, no bias \\
      \midrule
      2d Batch-normalization \\
      \midrule
      ReLU activation layer \\
      \midrule
      2d upsampling ($\times 2$) \\
      \midrule
      Conv2d $1 \times 1$, stride 1, padding 0, $64 > C$, with bias \\
      \midrule
      Spatial (nearest) interpolation $16 > 128$ \\
      \midrule
      Output $C \times 128 \times 128$ \\
      \bottomrule
\end{tabular}
\captionof{table}{Segmentation module}\label{tab:segmentation_module}
\end{table}

\begin{table}[h]
\centering
   \begin{tabular}{c}
   Classifier \\
   \toprule \midrule
      Input* $C \times 16 \times 16$ matrix \\
      \midrule
      Adaptive 2d average pooling $(16, 16) > (1, 1)$ \\
      \midrule
      Conv2d $1 \times 1$, stride 1, padding 0, $C > 8$, no bias \\
      \midrule
      2d Batch-normalization \\
      \midrule
      ReLU activation layer \\
      \midrule
      Conv2d $1 \times 1$, stride 1, padding 0, $8 > C$, no bias \\
      \midrule
      Output vector $\in \mathbb{R}^C$ \\
      \bottomrule
\end{tabular}
\captionof{table}{Classification module, * the input for the classifier is the segmentation output prior to spatial interpolation (see Table \ref{tab:segmentation_module}).} \label{tab:classification_module}
\end{table}

\section{Data preprocessing}
\label{apx:data_preprocess}

\subsection{Background removal}\label{apx:bg_removal}

\begin{figure}
     \centering
     \begin{subfigure}[b]{0.225\textwidth}
         \centering
         \includegraphics[width=\textwidth]{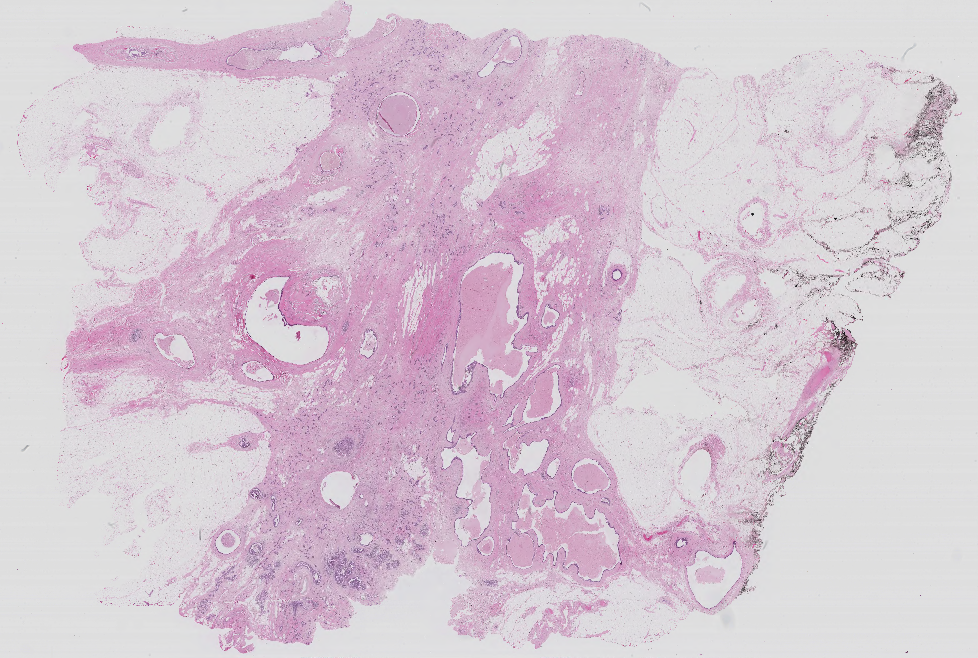}
     \end{subfigure}
     \hfill
     \begin{subfigure}[b]{0.225\textwidth}
         \centering
         \includegraphics[width=\textwidth]{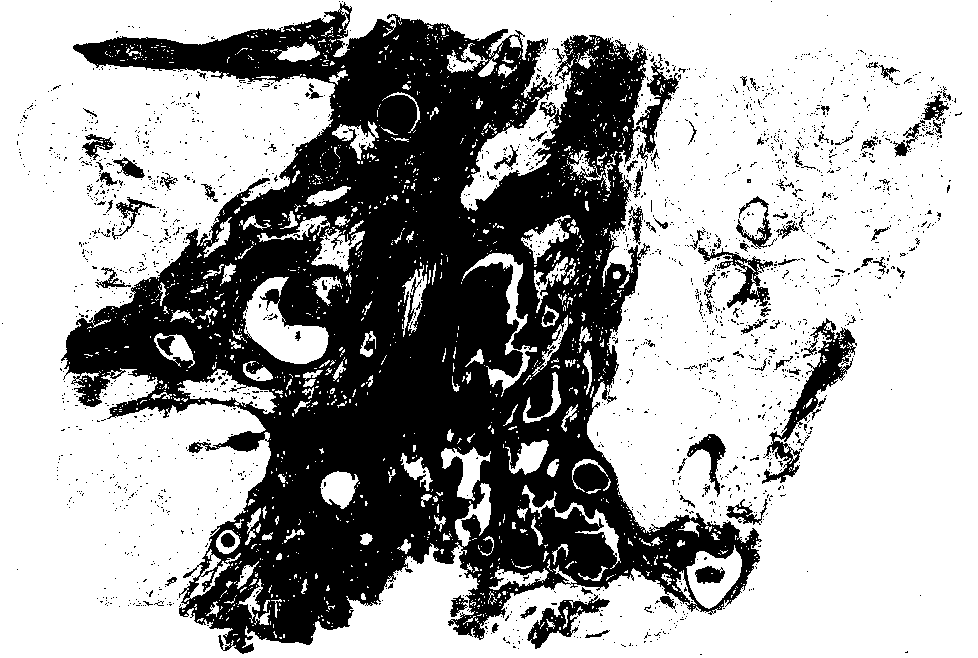}
     \end{subfigure}
     \begin{subfigure}[b]{0.225\textwidth}
         \centering
         \includegraphics[width=\textwidth]{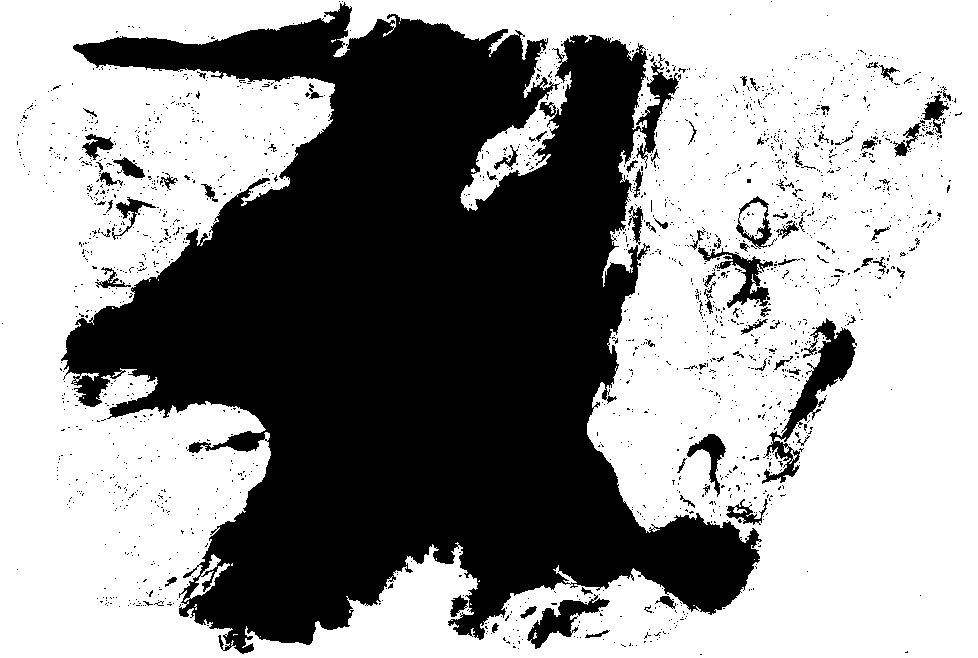}
     \end{subfigure}
     \hfill
     \begin{subfigure}[b]{0.225\textwidth}
         \centering
         \includegraphics[width=\textwidth]{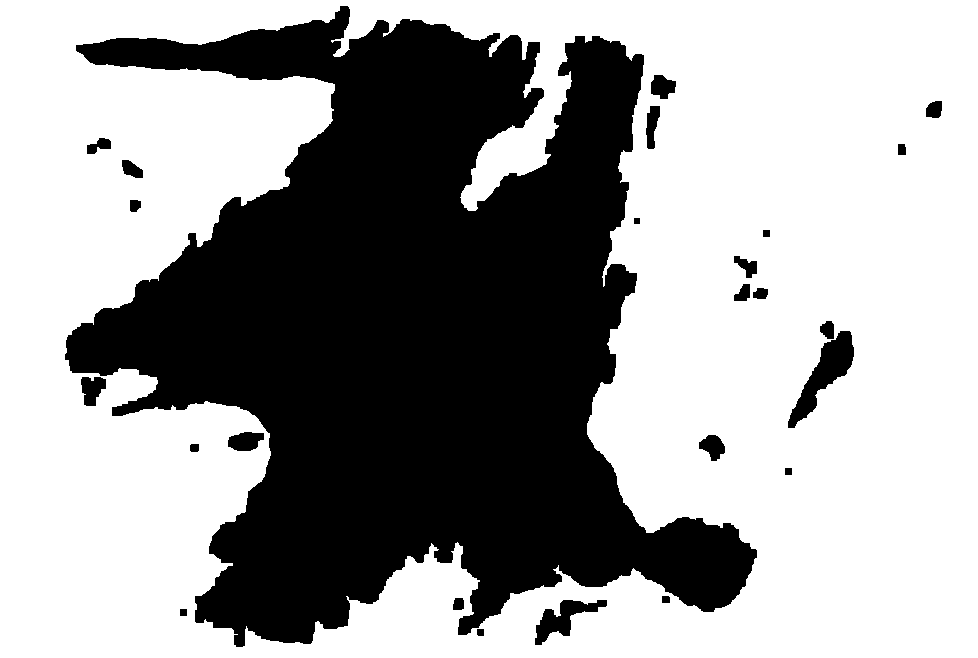}
     \end{subfigure}
        \caption{Background removal by HSV thresholding. The process is visualized from left to right, then top to bottom (row major order). }
        \label{fig:hsv_threshold}
\end{figure}

Given the segmentation mask of the WSI, one can build training data by extracting square patches of images and their corresponding ground truth mask as inputs to the training algorithm. A common method used is to slide a window over the WSI to get the patches. This method however, is very prone to cause imbalance issues. Insufficient training data and class imbalance are common problems in medical imaging. The problem is exacerbated in histopathology, and specifically WSIs, where image dimensions are extremely large, yet the labeled regions of interest can occupy less than 1\% of the image. Sliding a window over the whole image will likely generate many patches that only contain white background or stroma, and very few patches that contain non-background regions (e.g., invasive cancer). To alleviate this, we first generate a ``foreground mask" by applying a binary threshold of the saturation channel on the HSV color space image of the WSI. Specifically, pixels below 10\% of the maximum possible saturation value were considered as background. Then, the holes in the resulting mask are filled, as some of the ducts that are surrounded by stroma can be regions of interest (e.g., may contain ductal carcinoma in situ). Finally, opening operation is applied to remove the remaining regions that do not contain a consistently large foreground region (i.e., salt and pepper noise). The whole process is visualized in Fig. \ref{fig:hsv_threshold}. This foreground mask is used to discard patches that contain less than 75\% foreground, or white pixels shown in Fig. \ref{fig:hsv_threshold}. 

\subsection{Ground truth extraction}

\begin{figure}
     \centering
     \begin{subfigure}[b]{0.45\textwidth}
         \centering
         \includegraphics[width=\textwidth]{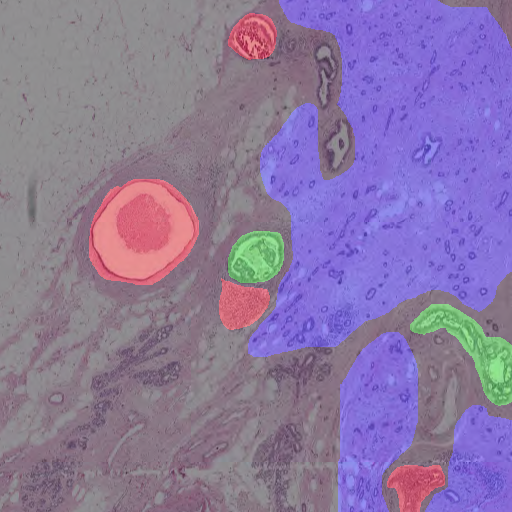}
         \caption{Centered extraction.}
         \label{fig:preprocess_centered}
     \end{subfigure}
     \hfill
     \begin{subfigure}[b]{0.45\textwidth}
         \centering
         \includegraphics[width=\textwidth]{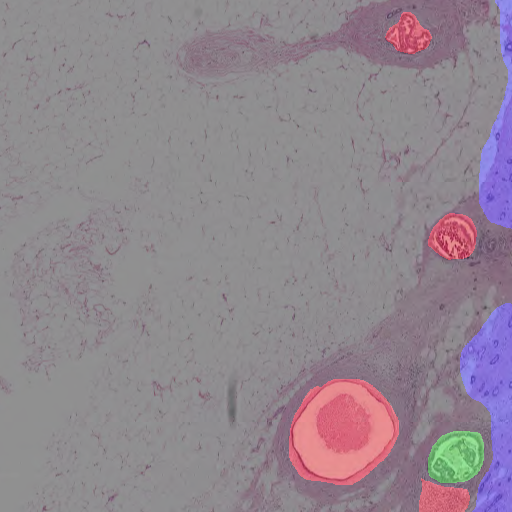}
         \caption{Tiled extraction}
         \label{fig:preprocess_tiled}
     \end{subfigure}
     \hfill
        \caption{Centered annotation extraction vs. windowed tiling.}
        \label{fig:centered_vs_window}
\end{figure}

In order to prevent a large class imbalance between foreground and background classes, we propose centered ground truth extraction. Unlike sliding a window with an arbitrary step size over the image, we use the center coordinates of each labeled connected component. Then each patch is extracted with endpoints $c_{x,y} \pm \frac{\Delta_{xy}}{2}$, where $c_{x,y}$ are $x$ and $y$ coordinates of the center of the patch, respectively, and $\Delta_{xy}$ is the square side length of the patch. In our experiments, we use a patch size $\Delta_{xy}$ 128, and we perform experiments in $1.25 X$ magnification of the WSI. The differences between sliding window approach and ours can be viewed in Fig. \ref{fig:centered_vs_window}, where we center on the green region (DCIS), whereas the sliding window is moved with a constant step size. Former is prone to extracting regions with vast background, and may only include small parts of regions of interest. As neural networks rely heavily on boundaries and context of the desired structure, it is hard for a network to learn from an incomplete picture. One may argue that this method will undersample the background regions, however even with this method, our dataset (number of total pixels in all patches) contains more than 90\% background. 

\begin{figure}[H]
     \centering
     \begin{subfigure}[b]{0.45\textwidth}
         \centering
         \includegraphics[width=\textwidth]{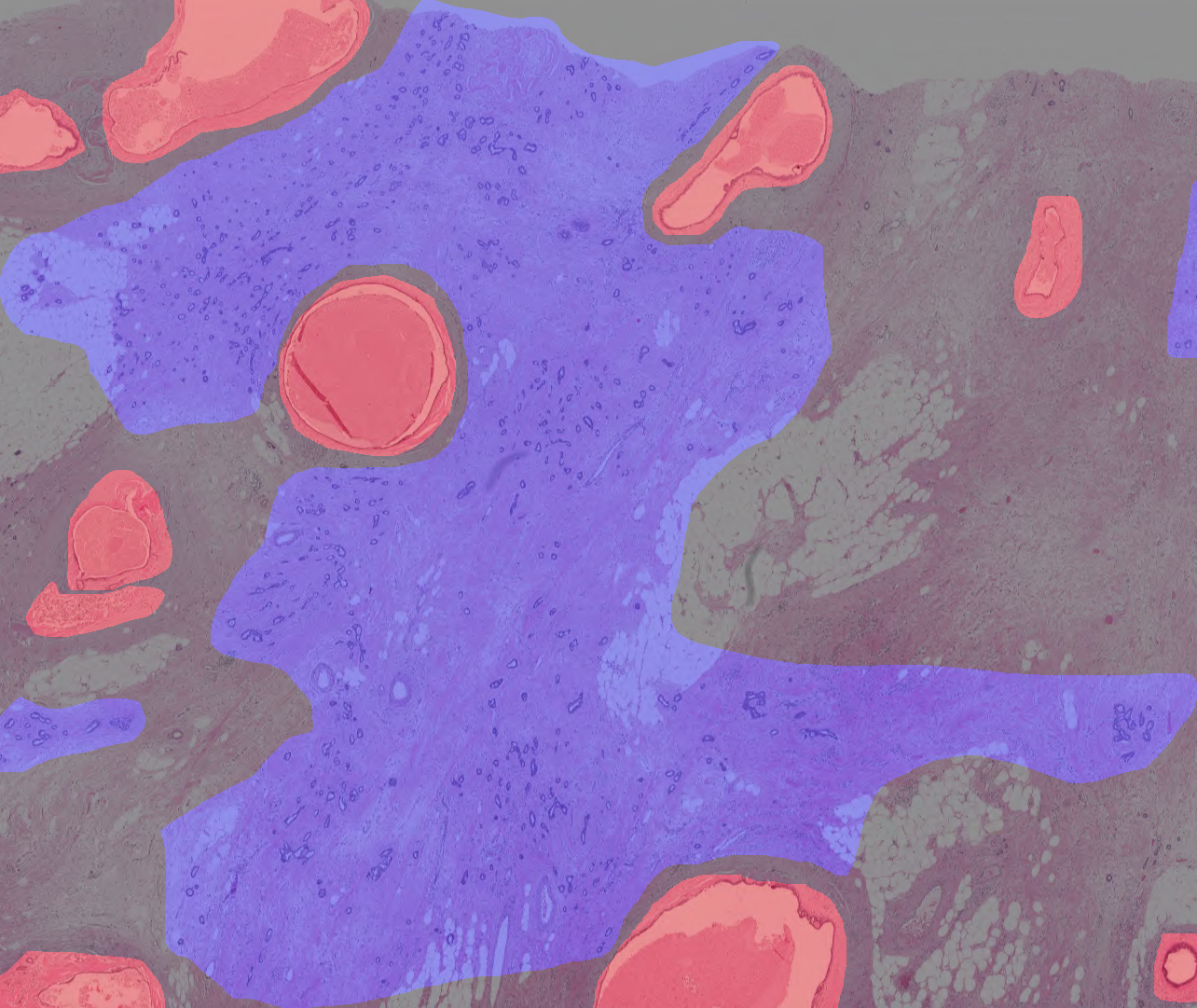}
         \caption{A large region of interest}
         \label{fig:k_means_large}
     \end{subfigure}
     \hfill
     \begin{subfigure}[b]{0.45\textwidth}
         \centering
         \includegraphics[width=\textwidth]{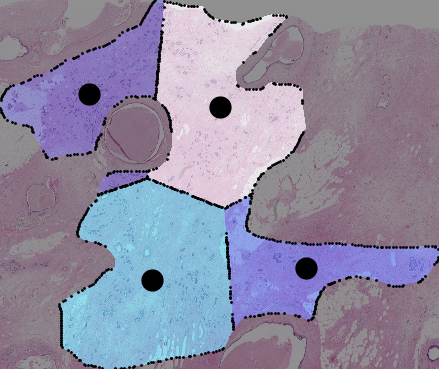}
         \caption{K-means based region splitting}
         \label{fig:k_means_large_split}
     \end{subfigure}
     \hfill
        \caption{Region splitting}
        \label{fig:k_means_region_split}
\end{figure}

In case of the region of interest is larger than our predefined patch size, we split the region into equal areas with the K-Means algorithm using the foreground $(x, y)$ pairs of coordinates as our inputs, and iterate the above procedure on each center. The number of ``clusters", or the centers, are determined based on the rule $\lceil 1 + \frac{Area_{region}}{\Delta_{xy}^2} \rceil$, where $\lceil \cdot \rceil$ is the ceiling function. The output of this process is visualized in Fig. \ref{fig:k_means_region_split}.

\onecolumn
\newpage

\section{Example segmentation results}

\begin{figure}[H]
     \begin{subfigure}[b]{1\textwidth}
         \centering
         \includegraphics[width=0.85\textwidth]{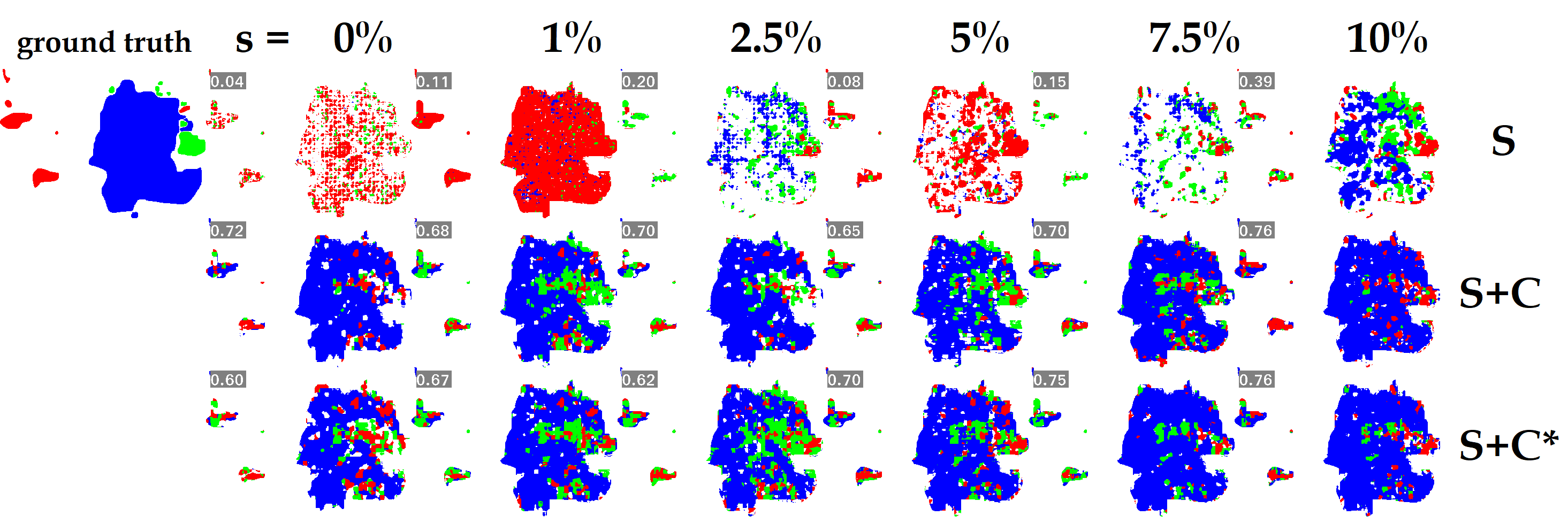}
         \label{fig:visual_samples_a02}
     \end{subfigure}
    %\hfill

     \begin{subfigure}[b]{1\textwidth}
         \centering
         \includegraphics[width=0.85\textwidth]{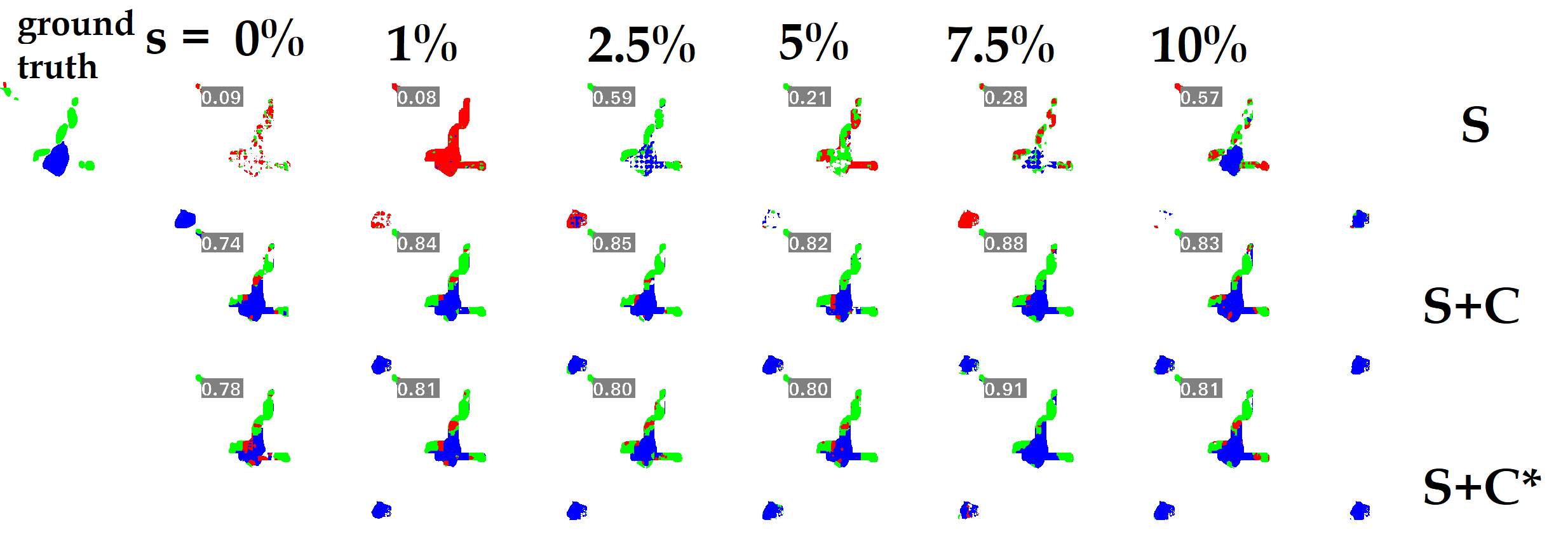}
         \label{fig:visual_samples_a03}
    \end{subfigure}
    %\hfill
    
     \begin{subfigure}[b]{1\textwidth}
         \centering
         \includegraphics[width=0.85\textwidth]{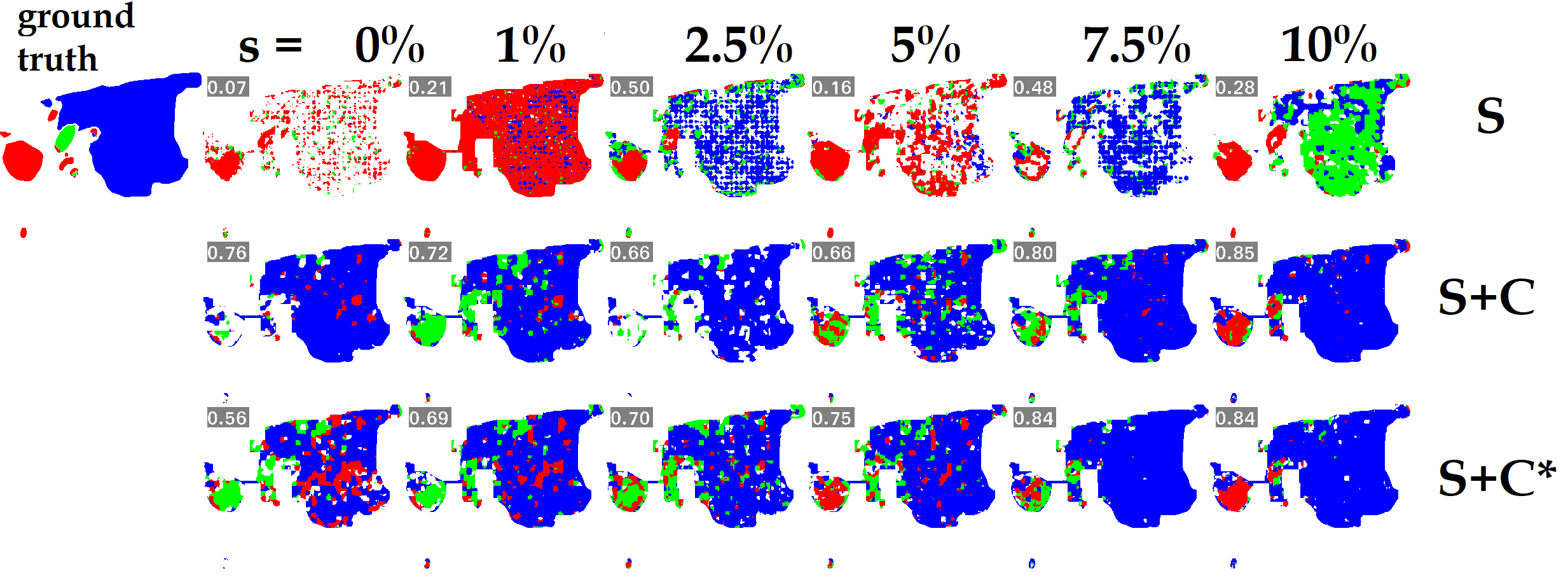}
         \label{fig:visual_samples_a05}
     \end{subfigure}
     \caption{Sample outputs from the ICIAR BACH 2018 dataset. Samples with $s=0, 1, 2.5, 5, 7.5, 10$ (left to right) for \textbf{S} (only $s\%$ of the segmentation patches are used), \textbf{S+C} ($s\%$ of the segmentation, and 100-s\% of the classification patches are used), and \textbf{S+C*} ($s\%$ of the segmentation, and 100\% of the classification patches are used) settings. The first column on the top row is the ground truth, and the top row is the setting \textbf{S}, middle row is \textbf{S+C}, whereas the bottom row is the setting \textbf{S+C*}. The numbers on the top left of each image is the accuracy for each sample output. }
     \label{fig:visual_samples_bach}
\end{figure}

\begin{figure*}
%     \begin{subfigure}[b]{1\textwidth}
%         \centering
%         \includegraphics[width=\textwidth]{figures/compare_performance/gleason2019/seg_outputs/slide003_core055_mask_jet.jpg}
%         \label{fig:visual_samples_a02}
%     \end{subfigure}

     \begin{subfigure}[b]{1\textwidth}
         \centering
         \includegraphics[width=0.85\textwidth]{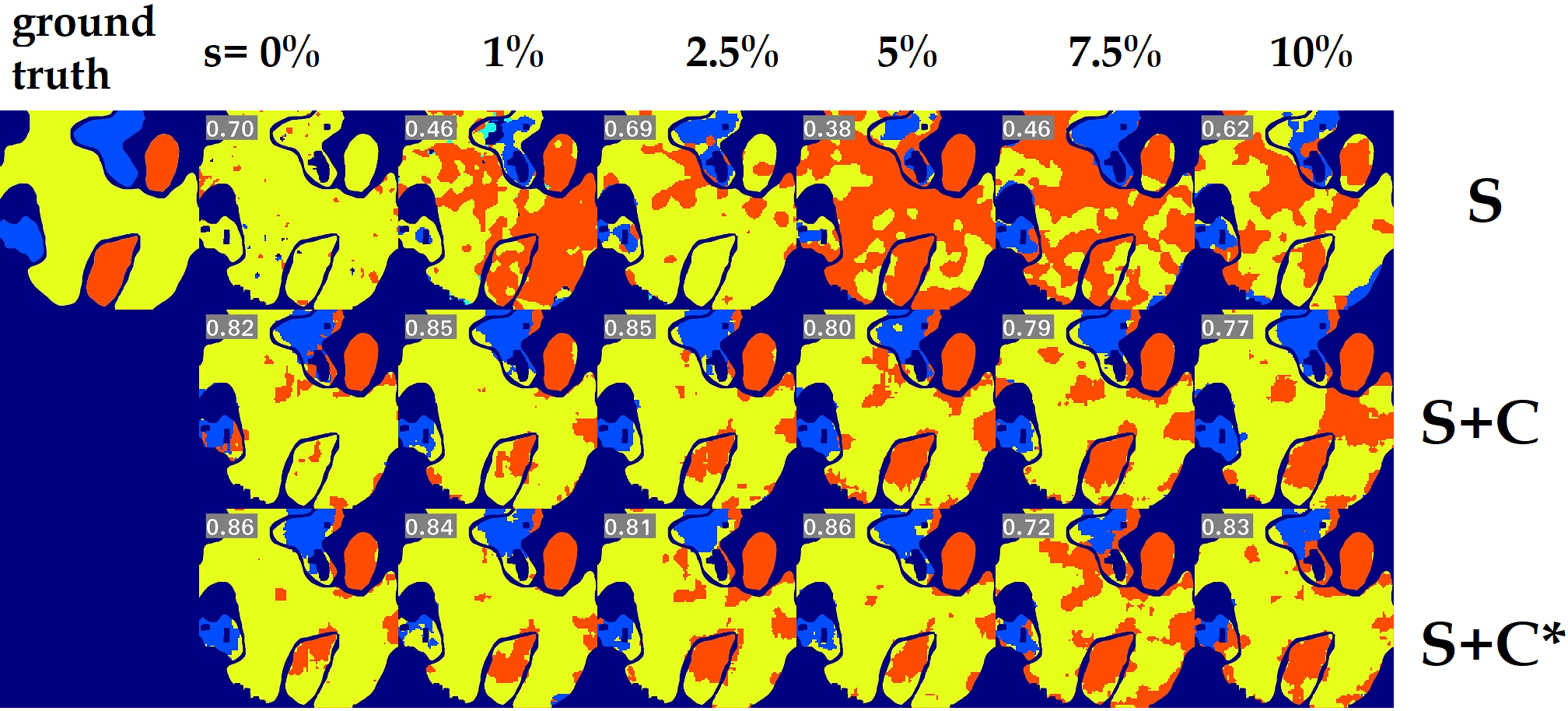}
         \label{fig:slide005_core012_mask_jet}
     \end{subfigure}
    \hfill

%     \begin{subfigure}[b]{1\textwidth}
%         \centering
%\includegraphics[width=\textwidth]{figures/compare_performance/gleason2019/seg_outputs/slide005_core021_mask_jet.jpg}
%         \label{fig:visual_samples_a04}
%     \end{subfigure}

     \begin{subfigure}[b]{1\textwidth}
         \centering
\includegraphics[width=0.85\textwidth]{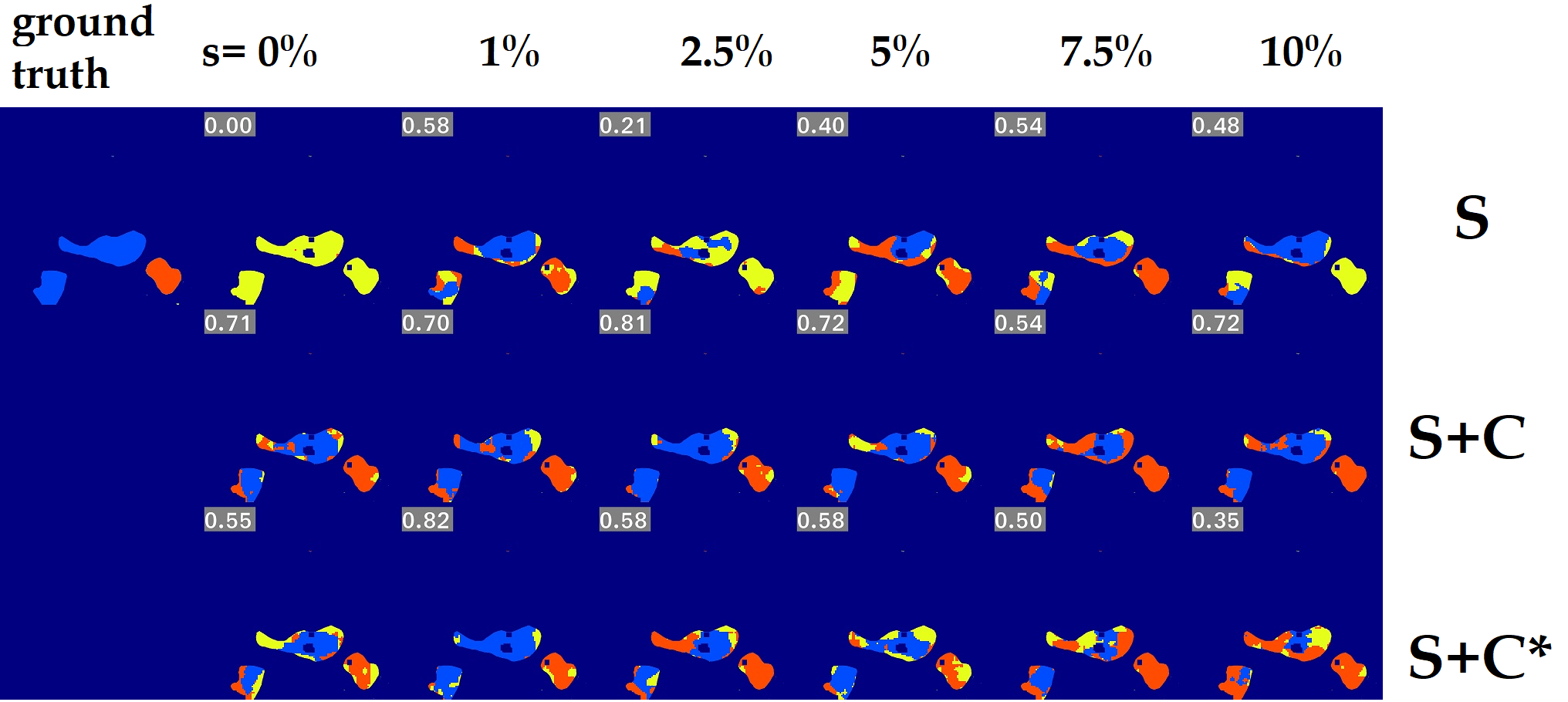}
         \label{fig:slide005_core051_mask_jet}
    \end{subfigure}
    \hfill

%    \end{figure*}
    
%    \begin{figure*}
%     \begin{subfigure}[b]{1\textwidth}
%         \centering
%\includegraphics[width=\textwidth]{figures/compare_performance/gleason2019/seg_outputs/slide005_core064_mask_jet.jpg}
%         \label{fig:visual_samples_a07}
%     \end{subfigure}
%     \vfill
%     \begin{subfigure}[b]{1\textwidth}
%         \centering
%\includegraphics[width=\textwidth]{figures/compare_performance/gleason2019/seg_outputs/slide006_core110_mask_jet.jpg}
%         \label{fig:visual_samples_a09}
%     \end{subfigure}
%     \vfill
     \begin{subfigure}[b]{1\textwidth}
         \centering
\includegraphics[width=0.85\textwidth]{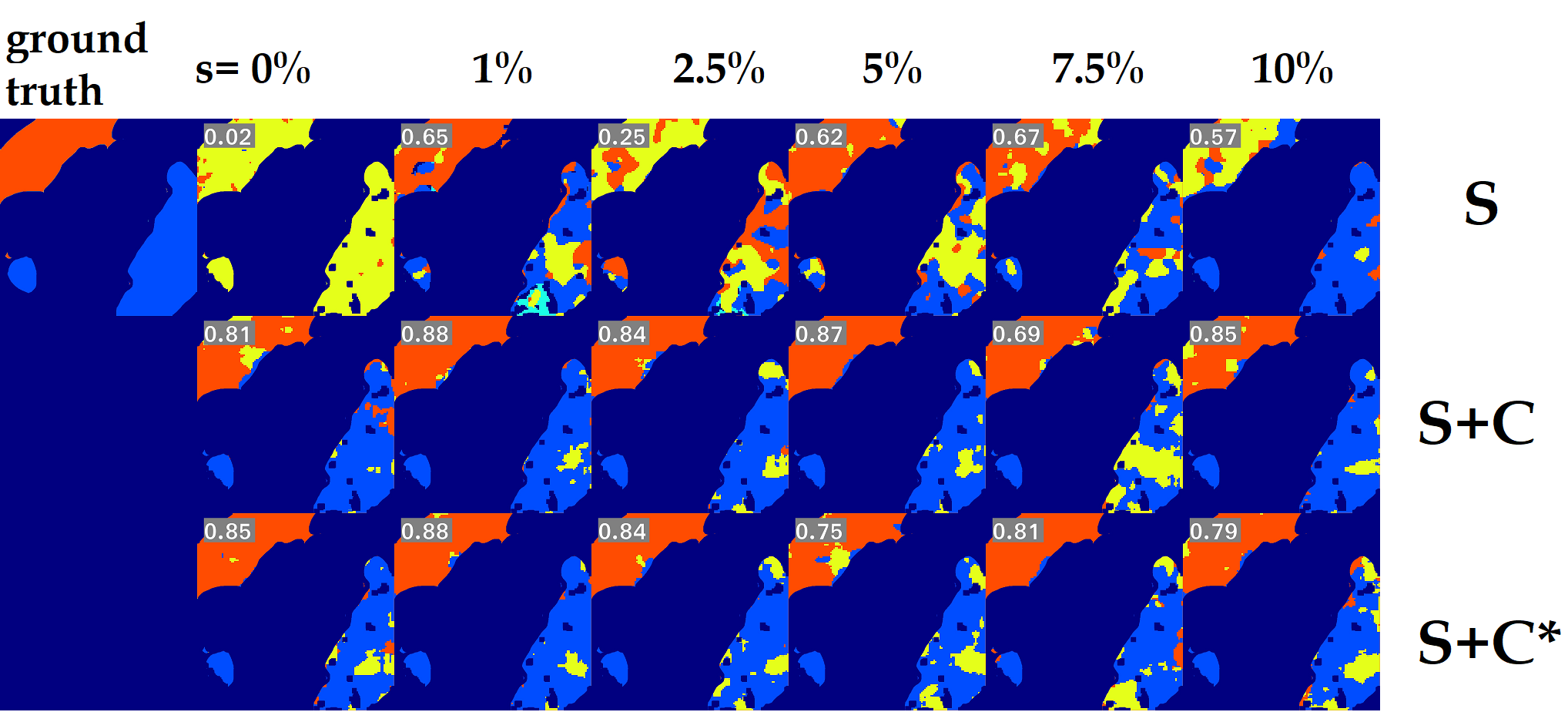}
         \label{fig:slide007_core146_mask_jet}
     \end{subfigure}
     \caption{Sample outputs from the Gleason2019 dataset. Samples with $s=0, 1, 2.5, 5, 7.5, 10$ (left to right) for \textbf{S} (only $s\%$ of the segmentation patches are used), \textbf{S+C} ($s\%$ of the segmentation, and 100-s\% of the classification patches are used), and \textbf{S+C*} ($s\%$ of the segmentation, and 100\% of the classification patches are used) settings. The first column on the top row is the ground truth, and the top row is the setting \textbf{S}, middle row is \textbf{S+C}, whereas the bottom row is the setting \textbf{S+C*}. The numbers on the top left of each image is the accuracy for each sample output.}
     \label{fig:visual_samples_gleason2019}
\end{figure*}

\begin{figure}
%     \begin{subfigure}[b]{1\textwidth}
%         \centering
%         \includegraphics[width=\textwidth]{figures/compare_performance/digestpath2019/seg_outputs/18-00530B-2019-05-07-23-56-22-lv1-11712-16122-7372-7686-mask-jet.jpg}
%         \label{fig:visual_samples_a02}
%     \end{subfigure}

     \begin{subfigure}[b]{1\textwidth}
         \centering
         \includegraphics[width=0.85\textwidth]{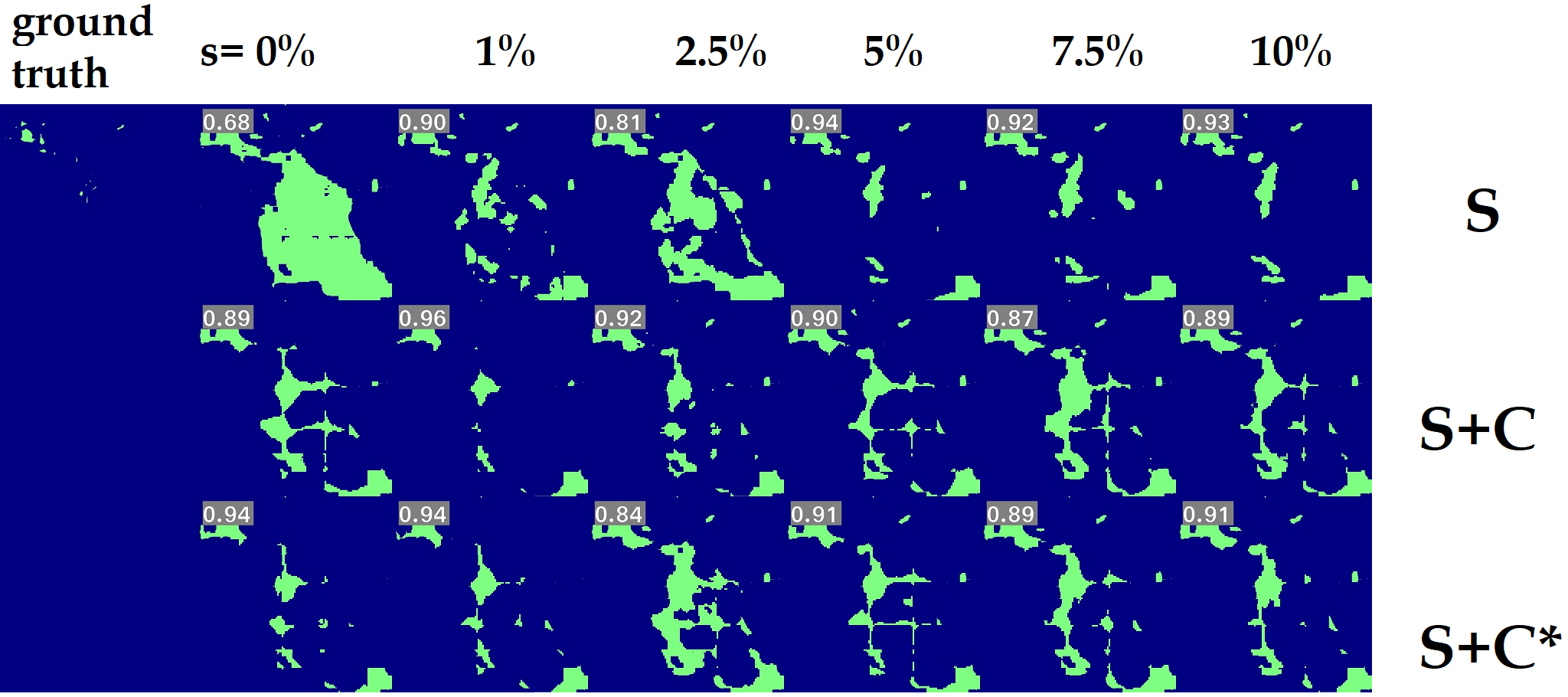}
         \label{fig:4484_5107_mask_jet}
     \end{subfigure}
    \hfill

%     \vfill
%     \begin{subfigure}[b]{1\textwidth}
%         \centering
%\includegraphics[width=\textwidth]{figures/compare_performance/digestpath2019/seg_outputs/18-14278B-2019-05-08-02-13-11-lv1-13964-11229-3779-5973-mask-jet.jpg}
%         \label{fig:visual_samples_a04}
%     \end{subfigure}
     \begin{subfigure}[b]{1\textwidth}
         \centering
         \includegraphics[width=0.85\textwidth]{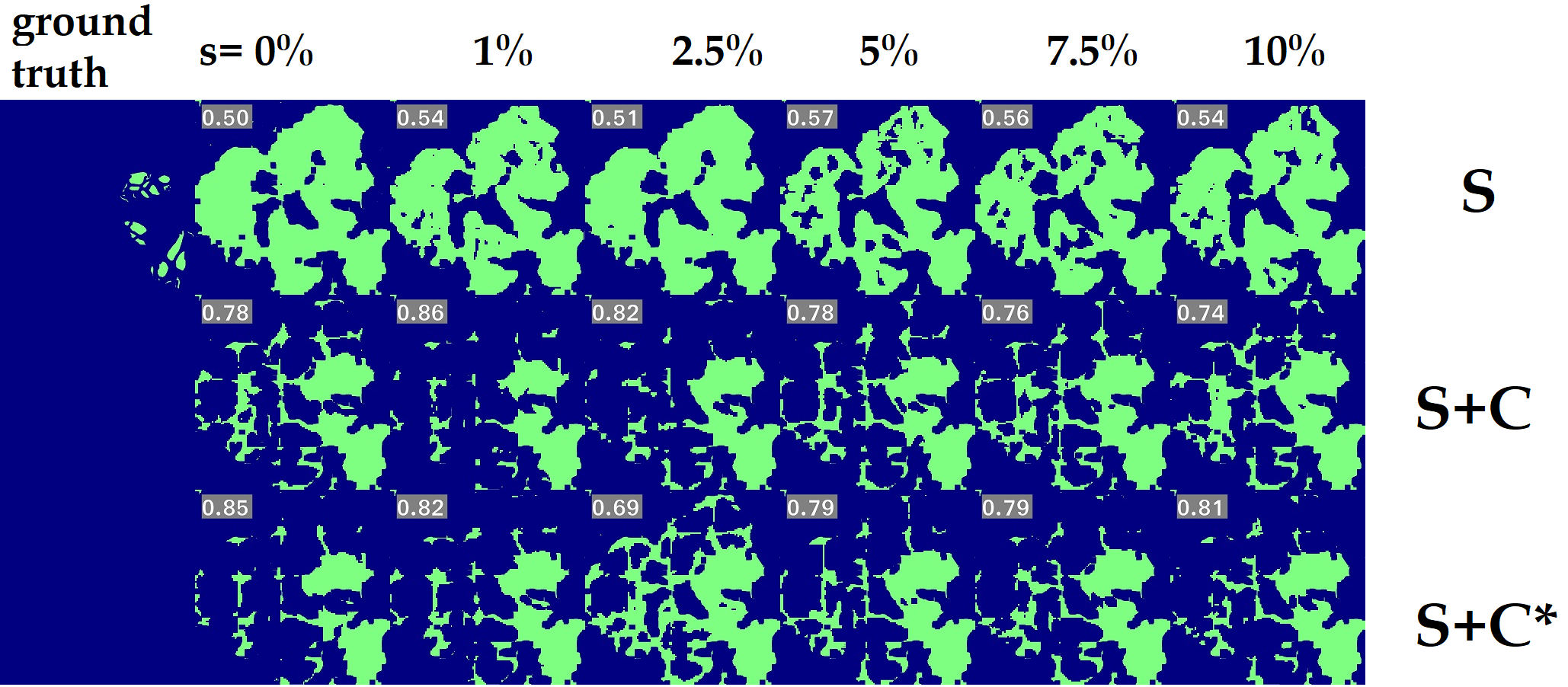}
         \label{fig:7406_7672_mask_jet}
     \end{subfigure}
    \hfill
    
%    \end{figure*}
    
%    \begin{figure*}\ContinuedFloat
     \begin{subfigure}[b]{1\textwidth}
         \centering
         \includegraphics[width=0.85\textwidth]{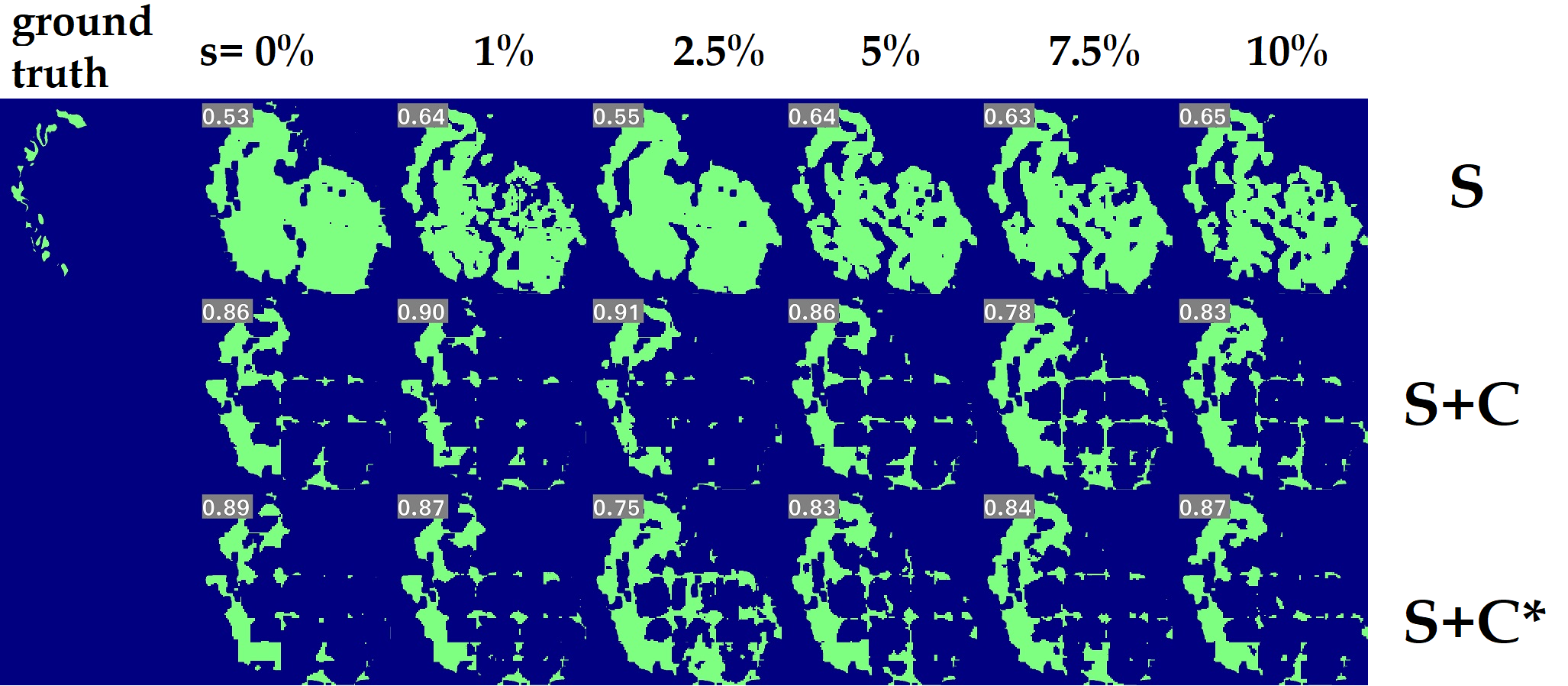}
         \label{fig:4857_4637_mask_jet}
     \end{subfigure}
     \caption{Sample outputs from the DigestPath2019 dataset. Samples with $s=0, 1, 2.5, 5, 7.5, 10$ (left to right) for \textbf{S} (only $s\%$ of the segmentation patches are used), \textbf{S+C} ($s\%$ of the segmentation, and 100-s\% of the classification patches are used), and \textbf{S+C*} ($s\%$ of the segmentation, and 100\% of the classification patches are used) settings. The first column on the top row is the ground truth, and the top row is the setting \textbf{S}, middle row is \textbf{S+C}, whereas the bottom row is the setting \textbf{S+C*}. The numbers on the top left of each image is the accuracy for each sample output.}
     \label{fig:visual_samples_digestpath2019}
\end{figure}

\newpage

\section{Numerical results}

\begin{table}[h]
\caption{Raw accuracy results for classification tasks when different combinations (modes) of training data are used. These results are raw or unnormalized, and they represent actual metric values as opposed to the relative performance with respect to the performance at $c=50\%$, unlike the presentation given in Fig. \ref{fig:classification_head_results}. The numbers in the top row represent the percentage of the used classification patches (c\%). For the \textbf{$S_2+C_2$} setting, 100-2c\% of segmentation and c\% of classification patches are used. For \textbf{$S_2+C^*_2$}, 100-2c\% of segmentation, and 50\% of classification patches are used. For the \textbf{$S^*_2+C_2$} setting, 100\% of the segmentation and c\% of classification patches are used.}
\centering
\begin{tabularx}{\linewidth}{X p{1cm}X | X X X X X X X X X X X X}
\toprule
Dataset & Mode & 0 & 1 & 2.5 & 5 & 7.5 & 10 & 15 & 20 & 25 & 30 & 40 & 50 \\
\midrule
ICIAR BACH 2018 & $S_2+C_2$ & 20.0 $\pm$ 7.9 &  34.4 $\pm$ 5.9 &  40.0 $\pm$ 10.2 &  48.0 $\pm$ 8.6 &  52.0 $\pm$ 6.2 &  62.4 $\pm$ 10.7 &  64.0 $\pm$ 6.1 &  67.2 $\pm$ 3.6 &  69.6 $\pm$ 1.3 &  72.0 $\pm$ 1.7 &  74.4 $\pm$ 2.5 &  80.0 $\pm$ 2.6  \\  
 & $S_2+C^*_2$ & 74.4 $\pm$ 1.9 &  72.0 $\pm$ 1.9 &  80.0 $\pm$ 1.4 &  72.8 $\pm$ 0.5 &  71.2 $\pm$ 1.9 &  72.0 $\pm$ 3.9 &  70.4 $\pm$ 6.0 &  67.2 $\pm$ 3.3 &  68.0 $\pm$ 4.4 &  70.4 $\pm$ 5.6 &  66.4 $\pm$ 2.9 &  65.6 $\pm$ 3.6  \\  
 & $S^*_2+C_2$ &  19.4 $\pm$ 0.0 &  25.1 $\pm$ 0.0 &  33.2 $\pm$ 0.0 &  25.3 $\pm$ 0.0 &  35.0 $\pm$ 0.0 &  39.9 $\pm$ 0.0 &  45.5 $\pm$ 0.0 &  53.6 $\pm$ 0.0 &  53.7 $\pm$ 0.0 &  57.0 $\pm$ 0.0 &  61.9 $\pm$ 0.0 &  68.4 $\pm$ 0.1  \\

\midrule
Gleason 2019 & $S_2+C_2$ & 12.1 $\pm$ 7.8 &  21.2 $\pm$ 6.0 &  26.6 $\pm$ 10.3 &  38.8 $\pm$ 8.9 &  34.3 $\pm$ 6.2 &  48.4 $\pm$ 10.1 &  51.8 $\pm$ 6.4 &  53.2 $\pm$ 3.4 &  56.1 $\pm$ 1.6 &  60.1 $\pm$ 1.7 &  59.2 $\pm$ 2.8 &  65.0 $\pm$ 1.7 \\  
 & $S_2+C^*_2$ & 58.9 $\pm$ 1.4 &  57.2 $\pm$ 1.4 &  65.0 $\pm$ 1.1 &  59.7 $\pm$ 0.7 &  59.3 $\pm$ 1.9 &  56.1 $\pm$ 4.0 &  56.0 $\pm$ 6.2 &  50.7 $\pm$ 3.6 &  52.4 $\pm$ 4.3 &  56.7 $\pm$ 6.0 &  49.5 $\pm$ 2.9 &  52.8 $\pm$ 4.1  \\  
 & $S^*_2+C_2$ & 11.8 $\pm$ 0.0 &  15.4 $\pm$ 0.0 &  22.3 $\pm$ 0.0 &  14.6 $\pm$ 0.0 &  26.0 $\pm$ 0.0 &  30.1 $\pm$ 0.0 &  33.2 $\pm$ 0.0 &  40.5 $\pm$ 0.0 &  39.8 $\pm$ 0.0 &  44.2 $\pm$ 0.0 &  48.1 $\pm$ 0.0 &  54.1 $\pm$ 0.1 \\

\midrule
Digest Path 2019 & $S_2+C_2$ &  50.7 $\pm$ 7.6 &  45.1 $\pm$ 5.3 &  32.2 $\pm$ 10.8 &  58.0 $\pm$ 8.8 &  43.0 $\pm$ 6.7 &  60.4 $\pm$ 10.3 &  61.7 $\pm$ 6.2 &  65.9 $\pm$ 3.6 &  71.6 $\pm$ 2.0 &  70.3 $\pm$ 2.2 &  77.6 $\pm$ 2.9 &  82.0 $\pm$ 2.5  \\  
& $S_2+C^*_2$ &  75.0 $\pm$ 1.1 &  73.6 $\pm$ 1.4 &  82.0 $\pm$ 1.9 &  76.9 $\pm$ 1.2 &  73.6 $\pm$ 2.7 &  73.3 $\pm$ 4.2 &  68.7 $\pm$ 5.1 &  63.4 $\pm$ 4.1 &  69.5 $\pm$ 4.5 &  69.7 $\pm$ 6.1 &  69.1 $\pm$ 2.8 &  63.8 $\pm$ 4.0  \\  
& $S^*_2+C_2$ & 49.1 $\pm$ 9.2 &  55.9 $\pm$ 1.2 &  59.7 $\pm$ 0.5 &  48.1 $\pm$ 1.0 &  55.9 $\pm$ 0.5 &  57.9 $\pm$ 0.2 &  62.1 $\pm$ 0.4 &  66.5 $\pm$ 0.4 &  62.7 $\pm$ 0.7 &  62.4 $\pm$ 0.2 &  66.5 $\pm$ 0.3 &  70.9 $\pm$ 0.2  \\

\bottomrule
\label{tab:classification_head_results_raw}\end{tabularx}
\end{table}

\begin{table}
\caption{Results for the segmentation task when different combinations (modes) of training data are used. These results are raw or unnormalized, and they represent actual metric values as opposed to the relative performance with respect to the performance at $s=100\%$, unlike the presentation given in Fig. \ref{fig:c_s_comparison}. The numbers in the top row represent the percentage of the used segmentation patches (s\%). For the \textbf{S}, s\% of segmentation and 0\% of the classification patches are used. For the \textbf{S+C}, s\% of segmentation and 100-s\% of the classification patches are used. For \textbf{S+C*}, s\% of segmentation, and 100\% of classification patches are used.}
\centering
\scriptsize
% >{\hsize=.85\hsize}
\begin{tabularx}{\linewidth}{>{\hsize=1.15\hsize}X X | >{\hsize=.85\hsize}X X X X X X X X X X X X X X X}
\toprule
Dataset & Mode & Metric & 0 & 1 & 2.5 & 5 & 7.5 & 10 & 15 & 20 & 25 & 30 & 40 & 50 & 75 & 100 \\

\midrule

ICIAR BACH 2018 & S & $F_{1_{micro}}$ & 26.4 $\pm$ 6.1 &  31.3 $\pm$ 6.5 &  34.3 $\pm$ 3.1 &  29.8 $\pm$ 4.4 &  33.4 $\pm$ 5.4 &  37.9 $\pm$ 2.1 &  34.9 $\pm$ 2.9 &  40.5 $\pm$ 1.2 &  38.8 $\pm$ 1.0 &  40.4 $\pm$ 1.5 &  40.8 $\pm$ 2.4 &  42.0 $\pm$ 1.7 &  42.9 $\pm$ 1.5 &  42.0 $\pm$ 0.5  \\  

& & $F_{1_{macro}}$ & 61.5 $\pm$ 8.8 &  67.2 $\pm$ 8.7 &  69.2 $\pm$ 4.2 &  67.4 $\pm$ 4.4 &  72.6 $\pm$ 5.5 &  76.5 $\pm$ 3.3 &  73.9 $\pm$ 3.5 &  78.8 $\pm$ 3.0 &  78.1 $\pm$ 1.8 &  79.5 $\pm$ 1.5 &  79.8 $\pm$ 1.7 &  80.4 $\pm$ 1.2 &  81.5 $\pm$ 1.7 &  80.8 $\pm$ 0.6 \\

 & S+C & $F_{1_{micro}}$ &  39.5 $\pm$ 2.6 &  39.9 $\pm$ 0.6 &  41.3 $\pm$ 0.8 &  40.7 $\pm$ 1.5 &  40.4 $\pm$ 1.9 &  41.4 $\pm$ 1.3 &  40.9 $\pm$ 1.1 &  41.1 $\pm$ 0.9 &  41.1 $\pm$ 1.5 &  40.3 $\pm$ 0.4 &  40.7 $\pm$ 1.4 &  40.9 $\pm$ 0.9 &  41.6 $\pm$ 1.2 &  42.1 $\pm$ 0.9  \\  

 & & $F_{1_{macro}}$ &  76.6 $\pm$ 2.8 &  77.7 $\pm$ 1.7 &  77.6 $\pm$ 1.0 &  75.9 $\pm$ 2.7 &  75.9 $\pm$ 2.3 &  76.8 $\pm$ 0.6 &  77.7 $\pm$ 1.8 &  77.1 $\pm$ 1.5 &  78.4 $\pm$ 1.8 &  77.1 $\pm$ 1.0 &  77.4 $\pm$ 2.3 &  78.9 $\pm$ 1.3 &  79.6 $\pm$ 0.2 &  80.9 $\pm$ 1.2  \\

 & S+C* & $F_{1_{micro}}$ &  40.4 $\pm$ 1.3 &  40.0 $\pm$ 0.5 &  39.6 $\pm$ 2.2 &  41.2 $\pm$ 0.9 &  41.2 $\pm$ 1.5 &  41.6 $\pm$ 1.1 &  40.4 $\pm$ 1.2 &  40.7 $\pm$ 0.8 &  41.3 $\pm$ 0.7 &  40.9 $\pm$ 1.1 &  40.8 $\pm$ 1.0 &  42.3 $\pm$ 0.9 &  42.5 $\pm$ 0.7 &  42.7 $\pm$ 0.3  \\  

 & & $F_{1_{macro}}$ &   77.1 $\pm$ 2.4 &  76.1 $\pm$ 1.0 &  75.5 $\pm$ 2.3 &  76.6 $\pm$ 1.6 &  76.7 $\pm$ 2.0 &  77.9 $\pm$ 1.8 &  75.7 $\pm$ 1.9 &  76.0 $\pm$ 1.1 &  77.9 $\pm$ 1.0 &  77.3 $\pm$ 2.3 &  78.2 $\pm$ 1.0 &  79.8 $\pm$ 0.7 &  80.3 $\pm$ 1.4 &  80.6 $\pm$ 1.2  \\

\midrule

Gleason 2019 & S & $F_{1_{micro}}$ & 18.5 $\pm$ 13.0 &  28.6 $\pm$ 9.5 &  26.5 $\pm$ 3.3 &  35.0 $\pm$ 2.6 &  36.2 $\pm$ 4.1 &  34.6 $\pm$ 3.8 &  37.2 $\pm$ 1.2 &  37.7 $\pm$ 3.0 &  39.9 $\pm$ 1.6 &  39.9 $\pm$ 1.6 &  39.4 $\pm$ 1.7 &  39.1 $\pm$ 1.1 &  40.0 $\pm$ 1.2 &  39.5 $\pm$ 0.9  \\  

& & $F_{1_{macro}}$ & 7.2 $\pm$ 4.0 &  10.6 $\pm$ 4.4 &  15.7 $\pm$ 1.8 &  20.0 $\pm$ 1.6 &  21.1 $\pm$ 3.0 &  20.2 $\pm$ 2.7 &  21.0 $\pm$ 1.2 &  21.5 $\pm$ 2.3 &  20.8 $\pm$ 5.9 &  20.7 $\pm$ 5.9 &  20.5 $\pm$ 5.8 &  22.8 $\pm$ 1.0 &  23.7 $\pm$ 1.1 &  23.4 $\pm$ 0.7  \\

& S+C & $F_{1_{micro}}$ & 40.3 $\pm$ 0.2 &  39.8 $\pm$ 0.4 &  40.5 $\pm$ 0.3 &  40.2 $\pm$ 0.1 &  40.0 $\pm$ 0.2 &  40.3 $\pm$ 0.2 &  40.1 $\pm$ 0.2 &  39.8 $\pm$ 0.3 &  40.8 $\pm$ 0.4 &  40.5 $\pm$ 0.5 &  40.6 $\pm$ 0.4 &  40.3 $\pm$ 0.6 &  41.4 $\pm$ 0.5 &  41.0 $\pm$ 0.6  \\  
& & $F_{1_{macro}}$ & 22.8 $\pm$ 0.4 &  22.2 $\pm$ 0.6 &  23.0 $\pm$ 0.2 &  22.4 $\pm$ 0.3 &  22.4 $\pm$ 0.5 &  22.8 $\pm$ 0.2 &  22.7 $\pm$ 0.3 &  22.7 $\pm$ 0.4 &  20.4 $\pm$ 5.3 &  20.4 $\pm$ 5.3 &  20.5 $\pm$ 5.4 &  23.2 $\pm$ 0.3 &  23.6 $\pm$ 0.6 &  23.5 $\pm$ 0.3  \\

& S+C* & $F_{1_{micro}}$ &  39.9 $\pm$ 0.7 &  40.2 $\pm$ 0.3 &  40.1 $\pm$ 0.6 &  40.2 $\pm$ 0.7 &  39.8 $\pm$ 0.4 &  40.3 $\pm$ 0.4 &  40.4 $\pm$ 0.4 &  40.7 $\pm$ 0.3 &  40.4 $\pm$ 0.5 &  40.9 $\pm$ 0.4 &  41.0 $\pm$ 0.4 &  40.7 $\pm$ 0.4 &  40.8 $\pm$ 0.3 &  40.8 $\pm$ 0.3  \\  

& & $F_{1_{macro}}$ &  22.7 $\pm$ 0.3 &  22.1 $\pm$ 0.8 &  22.7 $\pm$ 0.4 &  22.5 $\pm$ 0.7 &  22.6 $\pm$ 0.5 &  22.9 $\pm$ 0.1 &  23.0 $\pm$ 0.2 &  20.3 $\pm$ 5.3 &  20.1 $\pm$ 5.2 &  20.6 $\pm$ 5.4 &  23.4 $\pm$ 0.3 &  23.3 $\pm$ 0.2 &  23.5 $\pm$ 0.3 &  23.5 $\pm$ 0.3  \\

\midrule

Digest Path 2019 & S & $F_{1_{micro}}$ & 10.8 $\pm$ 9.1 &  21.2 $\pm$ 4.3 &  32.5 $\pm$ 3.8 &  36.8 $\pm$ 4.3 &  36.1 $\pm$ 3.9 &  34.5 $\pm$ 1.8 &  36.4 $\pm$ 2.0 &  35.7 $\pm$ 1.0 &  38.5 $\pm$ 1.7 &  37.3 $\pm$ 0.2 &  38.8 $\pm$ 1.1 &  38.7 $\pm$ 1.4 &  40.4 $\pm$ 1.2 &  39.5 $\pm$ 1.1  \\  

& & $F_{1_{macro}}$ & 4.0 $\pm$ 2.8 &  8.8 $\pm$ 2.0 &  18.5 $\pm$ 3.1 &  20.3 $\pm$ 2.2 &  20.3 $\pm$ 1.9 &  20.5 $\pm$ 1.3 &  20.1 $\pm$ 1.8 &  20.2 $\pm$ 0.8 &  22.1 $\pm$ 1.5 &  21.3 $\pm$ 0.2 &  22.5 $\pm$ 0.9 &  22.9 $\pm$ 1.2 &  24.0 $\pm$ 1.2 &  23.3 $\pm$ 0.8  \\

& S+C & $F_{1_{micro}}$ & 40.3 $\pm$ 0.5 &  40.1 $\pm$ 0.3 &  40.5 $\pm$ 0.2 &  40.3 $\pm$ 0.5 &  40.0 $\pm$ 0.3 &  40.5 $\pm$ 0.2 &  40.2 $\pm$ 0.5 &  39.9 $\pm$ 0.6 &  40.5 $\pm$ 0.4 &  40.3 $\pm$ 0.3 &  40.2 $\pm$ 0.3 &  40.4 $\pm$ 0.3 &  41.0 $\pm$ 1.3 &  41.9 $\pm$ 1.6  \\  
& & $F_{1_{macro}}$ & 22.8 $\pm$ 0.4 &  22.8 $\pm$ 0.3 &  22.7 $\pm$ 0.2 &  22.9 $\pm$ 0.3 &  22.9 $\pm$ 0.4 &  23.0 $\pm$ 0.2 &  22.6 $\pm$ 0.5 &  22.8 $\pm$ 0.6 &  22.9 $\pm$ 0.4 &  23.2 $\pm$ 0.3 &  22.9 $\pm$ 0.3 &  23.1 $\pm$ 0.5 &  23.7 $\pm$ 0.9 &  24.0 $\pm$ 1.4  \\

& S+C* & $F_{1_{macro}}$ &  39.6 $\pm$ 0.7 &  40.1 $\pm$ 0.5 &  40.6 $\pm$ 0.3 &  40.0 $\pm$ 0.5 &  40.3 $\pm$ 0.2 &  40.4 $\pm$ 0.3 &  40.5 $\pm$ 0.4 &  40.4 $\pm$ 0.3 &  40.2 $\pm$ 0.4 &  40.4 $\pm$ 0.5 &  40.8 $\pm$ 0.2 &  40.8 $\pm$ 0.9 &  41.3 $\pm$ 0.2 &  41.3 $\pm$ 0.2  \\  
& & $F_{1_{macro}}$ &  22.2 $\pm$ 0.8 &  22.9 $\pm$ 0.2 &  22.8 $\pm$ 0.4 &  22.6 $\pm$ 0.4 &  22.5 $\pm$ 0.3 &  23.0 $\pm$ 0.5 &  22.8 $\pm$ 0.3 &  23.0 $\pm$ 0.2 &  22.9 $\pm$ 0.2 &  23.3 $\pm$ 0.3 &  23.4 $\pm$ 0.3 &  23.3 $\pm$ 0.7 &  23.7 $\pm$ 0.2 &  23.7 $\pm$ 0.2  \\  

\bottomrule
\label{tab:segmentation_results_raw}\end{tabularx}
\end{table}

\end{document}